\documentclass[lettersize,journal]{IEEEtran}
\usepackage{amsmath,amsfonts}
\usepackage{algorithmic}
\usepackage{array}
\usepackage[caption=false,font=normalsize,labelfont=sf,textfont=sf]{subfig}
\usepackage{textcomp}
\usepackage{stfloats}
\usepackage{url}
\usepackage{hyperref}
\hypersetup{
    colorlinks=true,
    linkcolor=black,
    filecolor=black,      
    urlcolor=black,
    citecolor=black,
}
\usepackage{verbatim}
\usepackage{graphicx}
\def\BibTeX{{\rm B\kern-.05em{\sc i\kern-.025em b}\kern-.08em
    T\kern-.1667em\lower.7ex\hbox{E}\kern-.125emX}}
\usepackage{balance}

\usepackage{graphics} 
\graphicspath{ {./images/} }
\usepackage{epsfig} 
\usepackage{cite}
\usepackage{xcolor}

\usepackage{booktabs}
\usepackage{multirow}
\usepackage{amsmath}
\usepackage{amssymb}
\usepackage{mathtools}
\usepackage{siunitx}
\usepackage{booktabs,tabularx,subcaption,array}
\usepackage{makecell}
\newcommand{\RomanNumeralCaps}[1]{\MakeUppercase{\romannumeral #1}}
\usepackage{pifont}
\newcommand{\cmark}{\textcolor{green!70!black}{\ding{51}}} 
\newcommand{\xmark}{\textcolor{red!80!black}{\ding{55}}}   
\newcommand{\circnum}[1]{\ding{\numexpr171+#1\relax}}
\usepackage{threeparttable}
\usepackage{soul}
\usepackage{xcolor}
\sethlcolor{yellow}
\usepackage{placeins}

\soulregister\cite7
\soulregister\emph7
\soulregister\ref7
\soulregister\eqref7

\newcolumntype{P}[1]{>{\centering\arraybackslash}m{#1}}

\begin{document}
\title{BiPneu: Design and Control of a Bipolar-Pressure Pneumatic System for Soft Robots}
\author{Yu Mei, Xinyu Zhou, Vedant Naik, Alan Gao and Xiaobo Tan, \IEEEmembership{Fellow, IEEE}
\thanks{This research was supported by National Science Foundation CMMI 1940950, ECCS 2024649 and CNS 2237577.}
\thanks{Yu Mei, Xinyu Zhou, Vedant Naik, Alan Gao and Xiaobo Tan are with the Department of Electrical and Computer Engineering, Michigan State University, East Lansing, MI 48824, USA. 
(email: {\small meiyu1@msu.edu}; {\small zhouxi63@msu.edu}; {\small naikveda@msu.edu}; {\small akgaomail@gmail.com}; {\small xbtan@msu.edu}) }%
}

\maketitle

\begin{abstract}
Positive–negative pressure regulation is critical to soft robotic actuators, enabling large motion ranges and versatile actuation modes. However, achieving high-performance regulation across both pressure polarities remains challenging due to asymmetric inflation–deflation dynamics, valve nonlinearities, and switching-induced flow disturbances. This paper presents \emph{BiPneu}, a scalable and cost-efficient multi-channel bipolar-pressure pneumatic system for soft robots that enables wide-range, accurate, and responsive pressure regulation while providing seamless compatibility with high-level software ecosystems. A dual-mode sliding-mode controller (DM-SMC) with hysteresis-supervised mode selection is proposed based on a hybrid electro-pneumatic model. Extensive simulation and experiments demonstrate the superior performance of DM-SMC in tracking step and sinusoidal pressure references compared with both advanced model predictive controllers and well-tuned PID controllers. Experimental results show average absolute errors of 1.44 kPa in multi-step tests and 4.23 kPa in sinusoidal tracking, corresponding to reductions of 11.9\% and 35.6\% relative to PID control, along with improved control effort, valve switching rate, and transient response. Robustness of DM-SMC is further verified on a bellow actuator with pressure-dependent volume. Finally, BiPneu’s capability is demonstrated via two soft robotic examples, quick ball-maneuvering with a soft parallel manipulator and real-time finite element method (FEM)-based teleoperation of a soft bellows actuator.
\end{abstract}


\section{Introduction}
\IEEEPARstart{S}{oft} robots have attracted increasing attention for their inherent compliance and adaptability, enabling safe exploration in unstructured environments \cite{katzschmann2018exploration} and robust interaction with humans \cite{hu2020novel, mei2024simultaneous}. Among various actuation technologies for soft robots, pneumatic actuation is particularly appealing because it offers high power-to-weight ratio and fast response with low-cost hardware. More recently, a growing class of \emph{bipolar-pressure} (i.e., positive--negative) pneumatic actuators has emerged \cite{fang2022efficient, hu2020novel}. These actuators exploit both \emph{positive} and \emph{negative} pressures to realize bidirectional motion, larger workspace, and versatile actuation modes. Despite these benefits, achieving rapid, precise, and robust pressure control across bipolar pressure regimes is challenging due to asymmetric inflation/deflation dynamics and strong valve nonlinearities.

The soft robotics community has developed an extensive set of pneumatic actuation systems, which can be broadly categorized into three classes \cite{chen2024programmable}: volume control, continuous pressure control using proportional valves, and switching pressure control using solenoid on/off valves. Volume control uses syringes or piston pumps to vary the chamber volume and induce deformation \cite{fang2022efficient}. However, displacement sources often provide limited airflow and are typically bulky. For continuous pressure control, proportional valves enable accurate regulation by accepting continuous analog voltages, eliminating the need for high-frequency pulse-width modulation (PWM) control \cite{park2025modeling, caasenbrood2022desktop, young2021control}. These valves can be easily driven by microprocessor units, such as Raspberry Pi. Such platforms natively support standard network protocols (e.g., TCP/IP) and robotic middleware frameworks (e.g., ROS~2), enabling seamless integration with high-level control and robotic software ecosystems.  However, proportional valves are generally much larger and more expensive than solenoid on/off valves (often 2--5$\times$ in size and 3--20$\times$ in cost) \cite{chen2024programmable, young2021control}, limiting compact and scalable implementations for robots requiring many independently controlled channels. In contrast, switching-valve architectures regulate pressure using PWM-driven solenoid on/off valves, enabling relatively compact and cost-effective multi-channel implementation. Consequently, a variety of on/off--valve-based pneumatic systems have been developed for soft robots \cite{massoud2025enhancing, holland2014soft}. However, existing platforms primarily target positive-pressure regulation, with only a few supporting bipolar pressure operation \cite{pei2025programmable, chen2024programmable, zhang2022pneumatic, tian2023openpneu}. Among these bipolar pressure systems with on/off valves, some realize regulation using per-channel mini pumps \cite{pei2025programmable, zhang2022pneumatic, tian2023openpneu}, which increases size and complexity and often constrains the achievable pressure range and transient response. Another type of bipolar pressure systems with on/off valves rely on low-level microcontroller units (MCUs) for PWM generation, which can hinder integration with high-level software stacks. Overall, existing platforms rarely achieve bipolar-pressure regulation within a cost-effective on/off–valve–based architecture, while supporting integration with high-level robotic software ecosystems.

In addition to hardware design, achieving high-performance control in on/off--valve-based bipolar pressure systems is difficult. While existing bipolar pressure platforms typically adopt PID for pressure regulation \cite{chen2024programmable, zhang2022pneumatic}, inflation--deflation switching induces transient flow disturbances and asymmetric dynamics due to air compressibility and valve nonlinearities, often exciting nonlinear oscillatory pressure responses that render fixed-gain PID tuning difficult and insufficient \cite{massoud2025enhancing}. To address this challenge, prior work has explored enhanced strategies such as fuzzy piecewise PID \cite{pei2025programmable}, dual-loop PID tuned via evolutionary algorithms \cite{massoud2025enhancing}, and model-based reinforcement learning \cite{park2025modeling}. While model-based control offers an appealing alternative, most efforts on characterization and modeling of pneumatic platforms are limited to individual components\cite{zhong2021investigation}. System-level modeling and closed-loop regulation of pneumatic systems remain largely unexplored.

\begin{table*}[t]
\caption{Comparison of our work with the state of the art in bipolar pressure pneumatic systems.}
\label{tab:pneumatic_comparison}
\centering
\renewcommand{\arraystretch}{0.9}
\setlength{\tabcolsep}{4pt}

\begin{threeparttable}

\newcommand{\descw}{3.2cm} 

\begin{tabularx}{\textwidth}{P{\descw}XXXXXX}
\toprule
\multirow{2}{*}{\textbf{Description}}
& \multicolumn{6}{c}{\textbf{Characteristics of $n$ channels}} \\
\cmidrule(lr){2-7}
& Reference\cite{zhang2022pneumatic}
& Reference\cite{tian2023openpneu}
& Reference\cite{chen2024programmable}
& Reference\cite{pei2025programmable}
& Reference\cite{caasenbrood2022desktop}
& \textbf{Our work} \\
\midrule

Number of channels ($n$)
& 6
& 10
& 8
& $\leq$ 8
& $\leq$ 12
& 16 \\

Air source
& $n$ mini pumps
& $2n$ mini pumps
& 1 air pump \newline 1 vacuum pump
& $n$ mini pumps
& 1 air pump \newline 1 vacuum pump
& 1 air pump \newline 1 vacuum pump \\

Valves
& $3n$ on--off valves
& $n$ on--off valves
& $2n$ on--off valves
& $3n$ on--off valves
& $n$ proportional valves
& $2n$ on--off valves \\

Control system
& $n$ Arduino
& 1 PCB
& 1 MCU
& 1 PCB
& 1 Raspberry Pi
& 1 Raspberry Pi \\

Regulation range
& $-59$--$112$~kPa
& $-50$--$80$~kPa
& $-50$--$150$~kPa
& $-60$--$100$~kPa
& $-100$--$100$~kPa
& $-80$--$200$~kPa \newline (up to $300$~kPa) \\

\parbox[c]{\linewidth}{\centering
Control accuracy\\
(w.r.t. regulation range)
}
& $< 3$~kPa (1.7\%)
& $< 0.2$~kPa (0.2\%)
& $< 1$~kPa (0.5\%)
& $< 1.5$~kPa (1.0\%)
& $< 1$~kPa (0.5\%)
& $< 1.5$~kPa (0.5\%) \\

Response time\circnum{1}
& $> 1.5$~s
& $> 0.7$~s
& $< 0.2$~s
& $0.6$~s
& $0.06$--$0.27$~s
& $0.13$~s \\

Software ecosystems compatibility\circnum{2}
& \xmark
& \xmark
& \xmark
& \xmark
& \cmark
& \cmark \\

Size
& 30$\times$24$\times$11~cm
& 32$\times$23$\times$12~cm
& 24$\times$16$\times$11~cm
& 10$\times$10$\times$2$\times n$~cm
& 15$\times$13$\times$10~cm
& 25$\times$25$\times$19~cm \\

Cost per channel\circnum{3}
& Medium
& Low
& Low
& Low
& High
& Low \\
\bottomrule
\end{tabularx}

\begin{tablenotes}[flushleft]
\footnotesize
\item[] \circnum{1}~Measured using a fixed 20~mL air volume with a pressure step of $0 \rightarrow 100$~kPa;
\ \circnum{2}~Supports TCP/IP networking and ROS~2--based robotic middleware;
\item[] \circnum{3}~Cost categories: High $>$ \$500, Medium \$100--\$500, and Low $<$ \$100.
\end{tablenotes}
\end{threeparttable}
\vspace{-5mm}
\end{table*}

In this work, we introduce \textbf{BiPneu}, a scalable and cost-efficient multi-channel \textbf{bi}polar-pressure \textbf{pneu}matic system that enables accurate and responsive regulation of both positive and negative pressures over a wide operating range, while providing compatibility with high-level robotic software ecosystems. A comparison between BiPneu and representative state-of-the-art bipolar-pressure pneumatic systems is summarized in Table~\ref{tab:pneumatic_comparison}. In our preliminary work~\cite{mei2025modeling}, we developed a hybrid electro-pneumatic model and a mixed-integer nonlinear MPC (MI-NMPC) framework that embeds the discrete inflation and deflation mode into the predictive-control formulation to co-optimize mode scheduling and PWM inputs. In this paper we substantially extend that study by  building a BiPneu prototype, experimentally identifying the proposed model, and examining the computational limitations of MI-NMPC. We further develop a dual-mode sliding-mode controller (DM-SMC) with hysteresis-supervised mode selection, which achieves accurate and responsive bipolar-pressure regulation while maintaining lightweight computation suitable for real-time embedded implementation. We validate the proposed system through simulation and experiments; in particular, experimental results on multi-step and sinusoidal pressure tracking with both a fixed-volume load and a soft bellows actuator with varying internal volume demonstrate the superior performance of the proposed DM-SMC. Finally, we demonstrate the capability and utility of BiPneu in two representative soft-robot scenarios. These include a ball-manipulation platform that demands accurate and responsive pressure regulation for dynamic interactions, and real-time FEM-based teleoperation of a soft bellow actuator. These demonstrations highlight BiPneu’s amenability to interfacing with high-level software.

The remainder of the paper is organized as follows. The BiPneu system design and software architecture are presented in Section~\RomanNumeralCaps{2}. The switched-system model for BiPneu is developed in Section~\RomanNumeralCaps{3}, along with the design and analysis of the proposed DM-SMC. Simulation and experimental results are reported in Section~\RomanNumeralCaps{4} and \RomanNumeralCaps{5}, respectively, followed by the demonstrations of BiPneu for ball manipulation and real-time FEM-based teleoperation in Section~\RomanNumeralCaps{6}. Finally, concluding remarks are provided in  Section~\RomanNumeralCaps{7}.

\section{Bipolar-Pressure Pneumatic System Design}

\subsection{Hardware Design}
\begin{figure*}[t]
  \centering
  \includegraphics[width=1.0\textwidth]{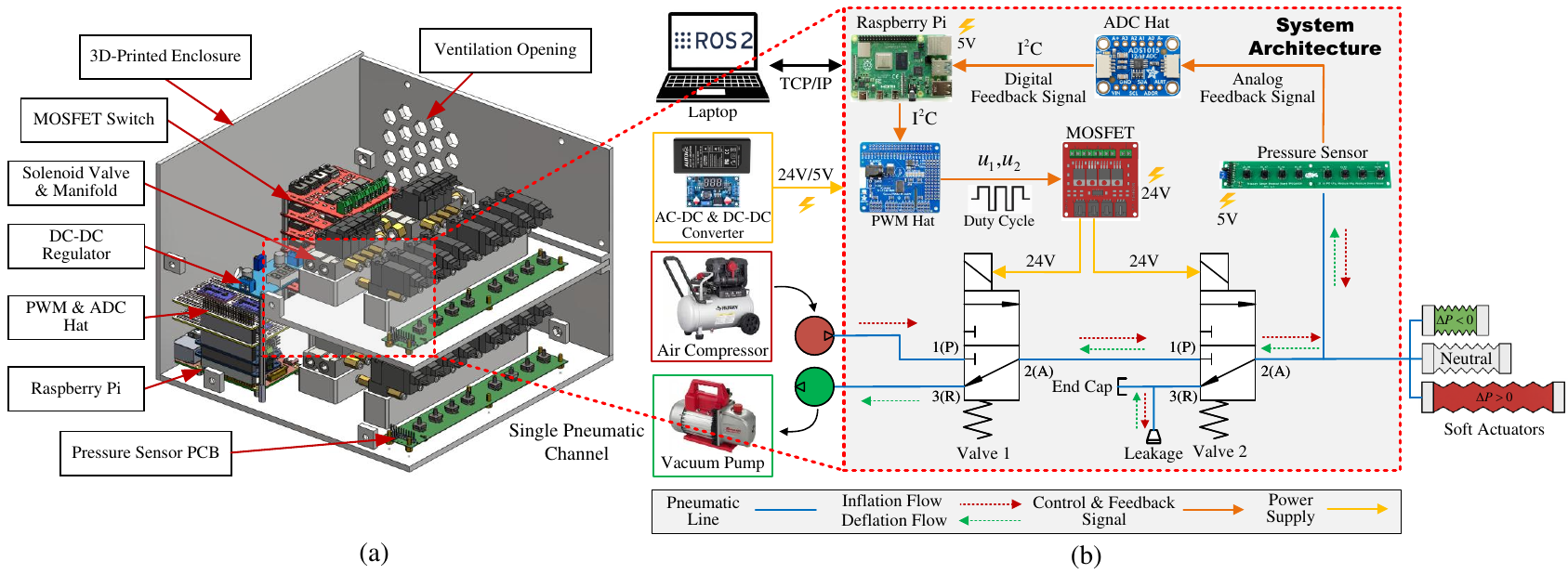}
  \vspace{-6mm}
  \caption{Overview of the BiPneu system. (a) Components and assembly layout of the BiPneu hardware. (b) Configuration of a single pneumatic channel, showing the pneumatic flow paths and the associated control/feedback signals and power connections.}
  \label{fig:system}
  \vspace{-5mm}
\end{figure*}

Fig.~\ref{fig:system} provides an overview of the BiPneu hardware architecture and the physical interconnections. As shown in Fig.~\ref{fig:system}(a), the entire system is packaged as a 16-channel unit within a 3D-printed enclosure (W$\times$L$\times$H: {25}$\times${25}$\times${19}~cm), comprising 16 identical per-channel regulation paths each with cascaded on/off valves, an embedded control subsystem, power electronics for valves, and pressure sensing boards. To support scalability, the valve-driving and pressure-sensing hardware is organized as two stacked 8-channel boards, with one 8-channel set mounted on the base board and a second 8-channel set on a sliding daughterboard. This stackable architecture supports additional channel expansion by adding sliding daughterboards. In practice, each added daughterboard contributes 8 more channels, while the control and power layers can be expanded modularly with additional PWM and MOSFET driver modules.

As shown in Fig.~\ref{fig:system}(b), which illustrates the configuration of a single pneumatic channel, BiPneu employs shared pressure sources, including an air compressor (Husky, No.~3301042) for inflation and a vacuum pump (Robinair, No.~15500) for deflation. Each channel comprises two cascaded 3/2 normally-closed solenoid on/off valves (SMC, VQ110U-5M-M5), driven by PWM inputs $u_1$ and $u_2$, respectively. For each valve, the outlet port (A) connects to the exhaust port (R) when de-energized, and to the inlet port (P) when energized. As shown in the lower half of Fig.~\ref{fig:system}(b), to direct the airflow, Valve~1 serves as a \emph{polarity selector}, routing the upstream node to either the positive-pressure line or the vacuum line, with the compressor and vacuum pump connected to its inlet (P) and exhaust (R) ports, respectively. The outlet port (A) of Valve~1 feeds into the inlet port (P) of Valve~2. Valve~2 serves as a \emph{delivery/hold} stage that meters flow to the soft actuator, and its exhaust port (R) is capped to isolate the channel when de-energized, thereby enabling pressure holding with small leakage-induced drift. Based on this pneumatic plumbing, the controller realizes per-channel inflation, deflation, and hold behaviors by selecting the pressure polarity via Valve~1 and modulating Valve~2 using PWM. 

At the system level, a Raspberry Pi 4B serves as the embedded controller and I/O hub, receiving high-level commands from a host computer over TCP/IP through Ethernet or Wi-Fi. Valve PWM commands are generated at 100~Hz and sent over I\textsuperscript{2}C to a PWM Hat (Adafruit PCA9685), which drives MOSFETs switching the 24~V solenoids. Pressure feedback is acquired through I\textsuperscript{2}C analog-to-digital converter (ADC) modules (Adafruit, ADS1015). To support 16-channel operation, the I\textsuperscript{2}C bus runs at 400~kHz with multi-threaded ADC polling. A custom printed circuit board (PCB) is developed for pressure monitoring and integrates eight pressure transducers (NXP, MPXHZ6400A). The supported pressure range of $-80$ to 300~kPa is determined by the operating range of the selected pressure sensor, and the PCB interfaces with the ADC Hats via headers to support scalable multi-channel integration. Regarding the power electronics, a 24~V AC/DC adapter (ALITOVE) powers the overall system, including the MOSFET-based solenoid switches, while an DC--DC converter (Seloky, LM2596) provides a regulated 5~V rail for the Raspberry Pi and the pressure-sensor board.

\subsection{Software Architecture}
\label{sec:software}
BiPneu implements a distributed ROS~2 server--client architecture that enables a high-throughput and versatile communication interface between high-level applications and embedded execution. The software architecture is described in Section~B and illustrated in Fig.~\ref{fig:supp_Software} in the Supplementary Material. On the client side, the embedded computer performs real-time pressure sensing, local control (e.g., PID or DM-SMC), and low-level valve actuation, while the server side generates desired pressure trajectories and supports system monitoring and data logging. Moreover, this ROS~2-based architecture offers broad software ecosystem compatibility. External high-level stacks, such as physics-based simulators (e.g., SOFA \cite{faure2012sofa}, MuJoCo \cite{todorov2012mujoco}, Isaac Sim \cite{isaacsim_github}), numerical toolchains (e.g., MATLAB/Simulink), and learning-based or vision--language--action policies running on distributed high-performance workstations or GPU servers, can compute pressure trajectories or task-level commands and publish them as standard ROS~2 messages without modifying the embedded execution layer.

\section{Dual-mode Sliding Mode Controller}

\subsection{Switched System Modeling of BiPneu}
The BiPneu system exhibits distinct airflow dynamics in inflation and deflation, and is therefore naturally modeled as a switched nonlinear system. We develop a physics-based switched model by combining a piecewise-continuous orifice-flow dynamics with mode-dependent flow-path switching. The detailed derivation of this switched model was presented in our preliminary work~\cite{mei2025modeling}, and is briefly summarized here. For each pneumatic channel, we take the outlet pressure $P_{\text{out}}(t)$ as the state, and use the PWM duty cycle $u_2(t)$ applied to Valve~2 as the continuous control input. For notational simplicity, we omit the explicit time argument $(t)$ and use $P \triangleq P_{\text{out}}$ and $u \triangleq {u_2}$ in the model below. The operating mode is represented by a binary variable $m\in\{0,1\}$ selected by Valve~1 through $u_1$, where $u_1=100\%$ sets $m=1$ corresponding to the inflation mode, and $u_1=0\%$ sets $m=0$ corresponding to the deflation mode. The pressure dynamics are modeled under an adiabatic assumption, as commonly used in prior work such as \cite{park2025modeling}. Under these definitions, the resulting dynamics can be expressed in the following hybrid form:
\begin{subequations}\label{eq:controlform}
\begin{equation}
\dot P = f(P) + g_m(P)\,\bar{x}(u), 
\end{equation}
where
\begin{equation}
f(P) = \frac{\gamma RT}{V}\,\big[ A_{ao}(P) - A_{oa}(P) \big],
\end{equation}
\begin{equation}
\resizebox{\columnwidth}{!}{$
\begin{aligned}
g_m(P) &= \frac{\gamma RT}{V}\,\Big(
    m\,[A_{po}(P) - A_{ao}(P) + A_{oa}(P)] \\
&\quad + (1-m)\,[-A_{on}(P) - A_{ao}(P) + A_{oa}(P)]
\Big),
\end{aligned}
$}
\end{equation}
\end{subequations}
where mode variable $m$ selects the active term in $g_m(P)$, and the nonlinear static function $\bar x(u)$ maps the PWM control input $u$ to the average spool position $\bar x$ ratio of Valve~2. This formulation models the net outlet flow as the superposition of the bidirectional main flow through Valve~1 and the leakage path through Valve~2. Accordingly, $A_{po}(P)$, $A_{on}(P)$, $A_{oa}(P)$, and $A_{ao}(P)$ denote the effective flow functions of the four branches, each defined from upstream to downstream, namely positive--outlet ($po$), outlet--negative ($on$), outlet--atmosphere ($oa$), and atmosphere--outlet ($ao$), given by:  
\begingroup
\setlength{\jot}{6pt}
\begin{subequations}\label{eq:Afunctions}
\begin{align}
A_{po}(P) &= P_{\text{pos}}\,C_{po}\,\rho_{\text{ref}}
             \sqrt{\tfrac{T_{\text{ref}}}{T}}\,
             \phi\!\left(\tfrac{P}{P_{\text{pos}}}\right), \\ 
A_{on}(P) &= P\,C_{on}\,\rho_{\text{ref}}
             \sqrt{\tfrac{T_{\text{ref}}}{T}}\,
             \phi\!\left(\tfrac{P_{\text{neg}}}{P}\right), \\ 
A_{oa}(P) &= P\,C_{oa}\,\rho_{\text{ref}}
             \sqrt{\tfrac{T_{\text{ref}}}{T}}\,
             \phi\!\left(\tfrac{P_{\text{atm}}}{P}\right), \\ 
A_{ao}(P) &= P_{\text{atm}}\,C_{ao}\,\rho_{\text{ref}}
             \sqrt{\tfrac{T_{\text{ref}}}{T}}\,
             \phi\!\left(\tfrac{P}{P_{\text{atm}}}\right).
\end{align}
\end{subequations}
\endgroup
where $\phi(r)$ is the shape factor defined by the piecewise function:
\begin{equation}
\phi(r) =
\begin{cases}
1, & r \leq b, \\[6pt]
\sqrt{\,1 - \left(\dfrac{r-b}{1-b}\right)^{2}}, & b < r < 1, \\[6pt]
0, & r \geq 1 ,
\end{cases}
\end{equation}
with $b$ denoting the critical pressure ratio. The symbols and parameters used in the orifice-flow model in \eqref{eq:controlform} and the branch-flow definitions in \eqref{eq:Afunctions} are summarized in Table~\ref{tab:pneu_symbols}.

\begin{table}[t]
  \caption{Parameters of the pneumatic mass–flow model.}
  \label{tab:pneu_symbols}
  \centering
  \renewcommand{\arraystretch}{1.05}
  \begin{tabularx}{\columnwidth}{@{} l X l l @{}}
    \toprule
    \textbf{Notion} & \textbf{Description} & \textbf{Unit} & \textbf{Value} \\
    \midrule
    $P_{\text{pos}}$ & Absolute pressure of the air compressor & \si{Pa} & \num{3.0e5} \\
    $P_{\text{neg}}$ & Absolute pressure of the vacuum pump & \si{Pa} & \num{1.0e4} \\
    $P_{\text{atm}}$ & Absolute pressure of the atmosphere & \si{Pa} & \num{1.01e5} \\
    $b$ & Critical pressure ratio & -- & \num{0.26} \\
    $C_{ud}$ & Sonic conductance from upstream to downstream & \si{m^{3}.s^{-1}.Pa^{-1}} & $^{\dagger}$ \\
    $\rho_{\text{ref}}$ & Gas density at reference conditions & \si{kg.m^{-3}} & \num{1.185} \\
    $T_{\text{ref}}$ & Reference temperature & \si{K} & \num{293.15} \\
    $T$ & Gas temperature & \si{K} & \num{293.15} \\
    $\bar{x}$ & Averaged spool position ratio & -- & $[0, 1]$ \\
    $u$ & PWM duty cycle & \% & $[0, 100]$ \\
    $\gamma$ & Heat capacity ratio & -- & \num{1.4} \\
    $R$ & Gas constant for air & \si{J.(kg\,K)^{-1}} & \num{287} \\
    $V$ & Outlet volume & \si{m^{3}} & \num{2.0e-5} \\
    \bottomrule
  \end{tabularx}
  
  \vspace{2pt}
  \footnotesize
  $^{\dagger}$~Parameter needs to be experimentally identified. 
  \vspace{-3mm}
\end{table}

\subsection{Dual-mode SMC Design}
\label{sec:DM-SMC}
Building on the dynamics (1)-(3), we develop a dual-mode sliding mode controller (DM-SMC) for BiPneu. The key idea is to design a mode-specific SMC law for the continuous dynamics within each operating mode and to switch the discrete mode with a hysteresis rule. Specifically, for a fixed mode $m\in\{0,1\}$, the systems has mode-dependent dynamics in Eq. \eqref{eq:controlform}, which enables SMC synthesis to compute the desired averaged spool-position ratio $\bar{x}$ and then obtain the PWM command $u$ via the inverse map of the calibrated $\bar{x}(u)$. 

To ensure reliable operation across inflation and deflation without rapid toggling near the setpoint, we introduce a hysteresis-supervised mode selector that updates $m$ based on the tracking error and holds the selected mode within a deadband. This hysteresis acts at the \emph{mode-selection} level and is distinct from the \emph{boundary layer} used later in the SMC law to mitigate chattering in the continuous control input. The mode selection strategy can be written as:
\begin{equation}\label{eq:mode_selection}
m =
\begin{cases}
1, & P \le P_{\mathrm{ref}} - h,\\
0, & P \ge P_{\mathrm{ref}} + h,\\
m_{\mathrm{prev}}, & \text{otherwise},
\end{cases}
\end{equation}
where $h>0$ is the hysteresis half-band, $P_{\mathrm{ref}}$ denotes the reference pressure, and $m_{\mathrm{prev}}\in\{0,1\}$ is the mode value from the previous control update. We define the tracking error:
\begin{equation}\label{eq:error}
e = P - P_{\mathrm{ref}},
\end{equation}
and the sliding surface $s\in\mathbb{R}$:
\begin{equation}\label{eq:sliding_surface}
s = \lambda e + k_I \int_0^t e(\tau)\,d\tau,
\end{equation}
where $\lambda>0$ is the sliding surface gain and $k_I\ge 0$ is the integral gain, which together tune the desired closed-loop dynamics on the sliding manifold. To enforce convergence to the sliding manifold while mitigating chattering, the reaching law is designed as:
\begin{equation}\label{eq:reaching_law}
\dot{s} = -\lambda s - \eta\,\mathrm{sat}\!\left(\frac{s}{\mu}\right),
\end{equation}
where $\eta>0$, and $\mu>0$ is the boundary-layer thickness of the high-slope saturation
$\mathrm{sat}(\cdot)$. For a fixed mode $m$, by differentiating Eq. \eqref{eq:sliding_surface}, one can obtain:
\begin{equation}\label{eq:sdot}
\dot{s} = \lambda(\dot P - \dot P_{\mathrm{ref}}) + k_I e
        = \lambda\!\left(f(P)+g_m(P)x-\dot P_{\mathrm{ref}}\right) + k_I e .
\end{equation}
Equating \eqref{eq:sdot} with the reaching law \eqref{eq:reaching_law} and solving for $x$ gives the control law:
\begin{equation}\label{eq:x_star}
\begin{aligned}
x^{\star} &=
\frac{-\,f(P) + \dot P_{\mathrm{ref}} - s
      - \dfrac{\eta}{\lambda}\,\mathrm{sat}\!\left(\dfrac{s}{\phi}\right)
      - \dfrac{k_I}{\lambda}\,e}
     {g_m(P)} , \\
u &= \bar{x}^{-1}\!\big(\mathrm{sat}_{[0,1]}(x^{\star})\big).
\end{aligned}
\end{equation}
where the saturation function $\mathrm{sat}_{[0,1]}(\cdot)$ clips to $[0,1]$ and $\bar{x}^{-1}(\cdot)$ is the inverse of the calibrated map $\bar{x}(u)$.

Overall, DM-SMC combines the hysteresis mode-selection rule in Eq.~\eqref{eq:mode_selection} with the mode-specific SMC law in Eq.~\eqref{eq:x_star} to coordinate discrete switching and continuous valve actuation. A conceptual illustration of the DM-SMC mechanism is provided in Fig. \ref{fig:supp_dm_smc_concept} in Section~C of the Supplementary Material.

\section{Simulation Results}
The proposed controller is first evaluated in simulation to isolate closed-loop dynamics from unmodeled hardware effects. The simulation is implemented in Python using the identified system model, with the same sampling rate and input constraints as the physical platform. White Gaussian noise $\mathcal{N}(0,0.5^2)$~kPa is added to the feedback measurement to emulate sensor noise. A constant load volume of 20~mL is assumed, matching the fixed-volume load in Fig.~\ref{fig:setup}(a). 

We evaluate the proposed DM-SMC alongside PID, nonlinear model predictive control (NMPC), and mixed-integer nonlinear model predictive control (MI-NMPC) under both step and sinusoidal pressure-tracking tasks. All controllers share the same hysteresis-supervised mode-selection rule in Eq.~\eqref{eq:mode_selection}. PID employs two mode-dependent controllers with separate gain sets; NMPC solves a nonlinear optimal control problem for the continuous PWM input under the hysteresis-selected mode, which is held fixed over the prediction horizon; and MI-NMPC jointly optimizes the discrete mode and PWM inputs within a predictive-control framework. The controller gains and weights are carefully tuned to achieve their best tracking performance. For the step response, the reference consists of multiple pressure stages
$(0,\,50,\,100,\,150,\,200,\,150,\,100,\,50,\,0,\,-40,\,-80,\,-40,\,0)\,$ kPa, each step holding for $5.0\,\text{s}$. For the sinusoidal case, the reference is defined as $P_{\text{ref}}(t)=50\sin(2\pi f t)$~kPa. Each controller is tested at three frequencies $f\in\{0.1,\,0.25,\,0.5\}$~Hz, to assess tracking performance from slow to fast references.

\begin{figure}[h]
\centerline{\includegraphics[scale=0.34]{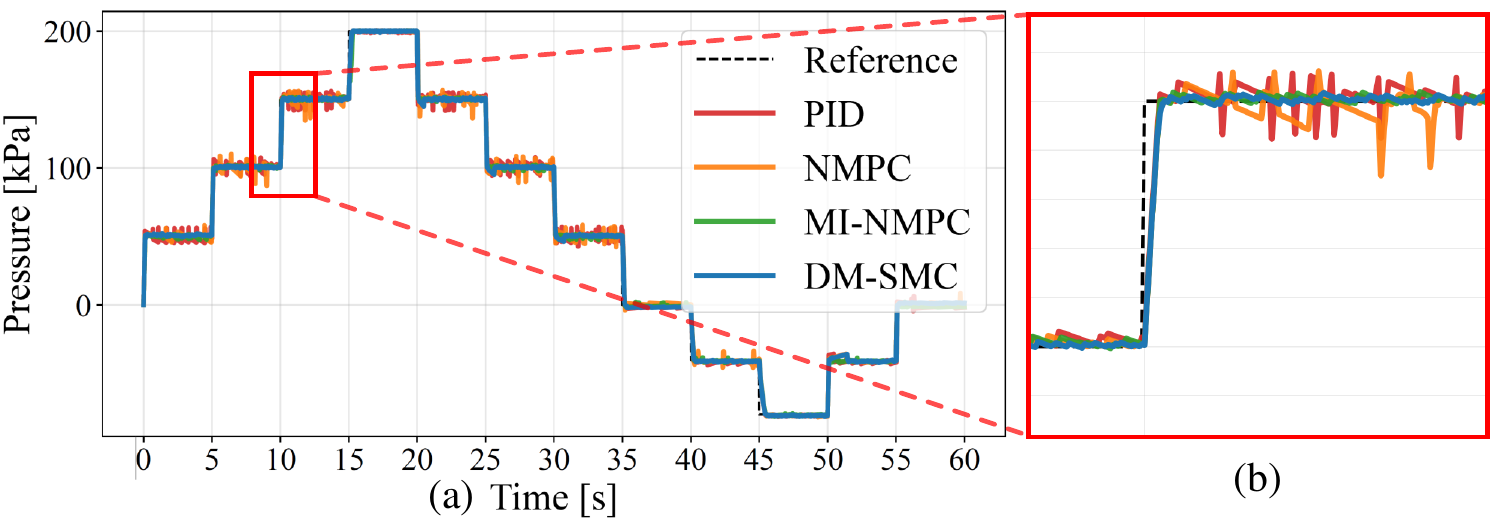}}
\vspace{-3mm}
\caption{Simulation results of multi-step reference-tracking performance for different controllers. (b) Zoomed-in view.}
\label{fig:Sim_Step}
\vspace{-5mm}
\end{figure}

\begin{figure}[h]
\centerline{\includegraphics[scale=0.26]{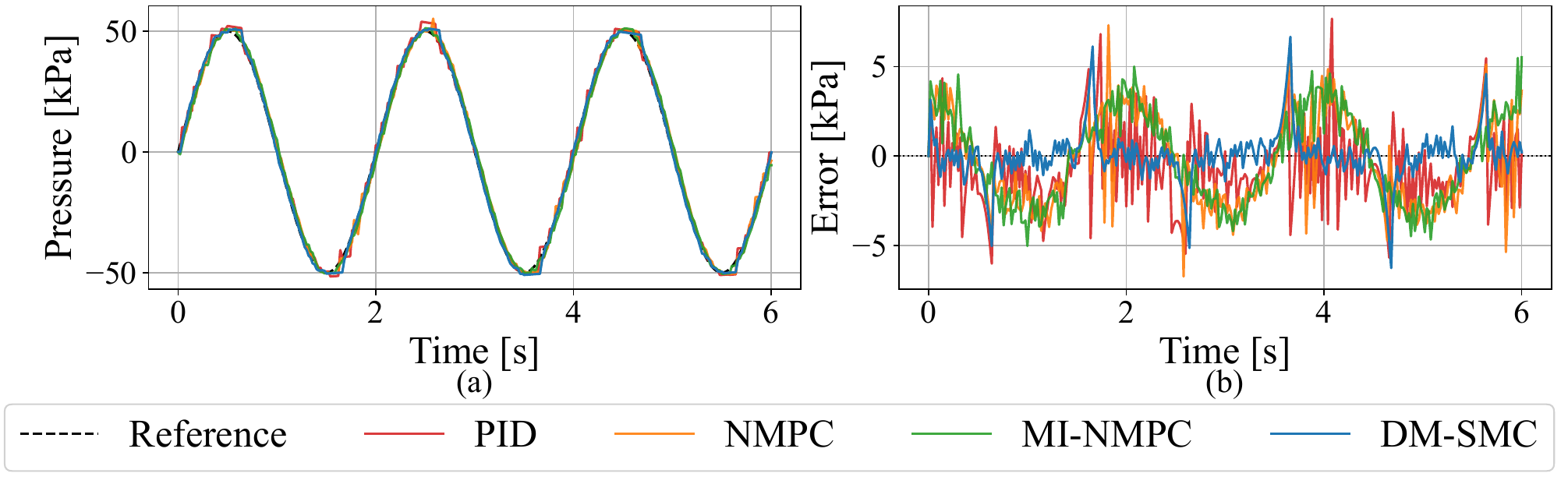}}
\vspace{-1mm}
\caption{Simulation results of sinusoidal reference-tracking at 0.5 Hz for four controllers: (a) pressure trajectories of the reference and controllers; (b) corresponding tracking errors.}
\label{fig:Sim_Sine}
\vspace{-5mm}
\end{figure}

The simulation results are shown in Fig.~\ref{fig:Sim_Step} and Fig.~\ref{fig:Sim_Sine}, where Fig.~\ref{fig:Sim_Sine} presents the representative 0.5~Hz case for brevity, and other sinusoidal tracking responses are reported in Fig.~\ref{fig:supp_Sim_Sine}. While all controllers are able to track the reference trajectories, they differ substantially in tracking accuracy, control effort, mode-switching behavior, and computational cost. Quantitative comparisons are summarized in Tables~\ref{tab:Sim_Step} and~\ref{tab:Sim_Sine}. For step tracking, performance metrics include the steady-state error $e_{ss}$, average absolute error (AE), stage-averaged integral of time-weighted absolute error (ITAE), stage-averaged PWM energy (PWM-E, defined as $\int |u(t)|dt$), stage-averaged number of mode switches, and the average computation time per control update (CT). Here, ``stage-averaged'' indicates that each metric is first computed over the analysis window of each individual pressure stage in the multi-step reference, and then averaged across all stages. For the sinusoidal-tracking case, the average absolute error and control-effort metrics are period-averaged, i.e., evaluated over each sinusoidal period and then averaged across periods. In addition, the maximum tracking error ($\max |e|$) is reported to characterize peak deviation under dynamic excitation. Computation time is reported for both a host PC (AMD~Ryzen~9~9900X CPU, 32~GB RAM) and the embedded Raspberry~Pi~4B. As shown in Table~\ref{tab:Sim_Step}, MI-NMPC achieves strong tracking with reduced control effort by co-optimizing mode scheduling and PWM inputs, but its computational cost is prohibitive for real-time embedded implementation. In contrast, the proposed DM-SMC provides more accurate tracking than PID and NMPC while remaining computationally lightweight, requiring only 0.692~ms on the embedded system. In the sinusoidal case, DM-SMC even outperforms MI-NMPC in most metrics, as shown in the Table~\ref{tab:Sim_Sine}. Its significantly reduced number of mode switches not only extends the lifetime of the solenoid valves but also indicates smoother control behavior, due to less overshoot and better suppression of unnecessary inflation/deflation switching. A plausible explanation is that sinusoidal tracking demands frequent mode decisions near zero crossings, where noise and solver tolerances can induce suboptimal switching and phase lag in MI-NMPC, whereas DM-SMC suppresses noise-induced switching through hysteresis-supervised sliding regulation and maintains tight tracking.

\begin{table}[h]
    \centering
    \caption{Multi-step reference-tracking performance in simulation: PID, NMPC, MI-NMPC and DM-SMC.} 
    \label{tab:Sim_Step}  
    \renewcommand{\arraystretch}{1.05}%
    \resizebox{1.0\columnwidth}{!}{%
    \begin{tabular}{@{}c@{} >{\raggedright\arraybackslash}p{2.5cm} | cccc@{}}
        \hline
        & \multicolumn{1}{c|}{Metrics} & PID & NMPC & MI-NMPC & DM-SMC \\
        \hline
        & $e_{\mathrm{ss}}$ [kPa]      & 1.67   & 1.27   & \textbf{0.73}    & 0.80    \\
        & AE [kPa]                     & 2.48   & 1.91   & \textbf{1.49}    & 1.59    \\
        & ITAE [kPa$\cdot$s$^2$]       & 22.15  & 15.00  & \textbf{9.58}    & 10.79   \\
        & PWM-E [\%$\cdot$s]           & 145.60 & 140.48 & \textbf{135.58}  & 137.11  \\
        & Switches                     & 9.08   & 6.08   & 9.08            & \textbf{0.83}    \\
        & CT (PC) [ms]                 & \textbf{0.002}   & 21.528 & 355.534          & 0.060   \\
        & CT (Embedded) [ms]           & \textbf{0.018}   & 270.819 & 3595.999        & 0.692   \\
        \hline
    \end{tabular}%
    }
    \begin{flushleft}
        \footnotesize \textbf{Bold} numbers represent the best results among all controllers for each metric.
    \end{flushleft}
    \vspace{-6mm}
\end{table}

\begin{table}[h]
    \centering
    \caption{Sinusoidal reference tracking performance in simulation: PID, NMPC, MI-NMPC, and DM-SMC.}
    \label{tab:Sim_Sine}  
    \vspace{-1mm}
    \renewcommand{\arraystretch}{1.05}%
    \resizebox{\columnwidth}{!}{%
    \begin{tabular}{@{}%
        >{\centering\arraybackslash}p{0.8cm}  %
        >{\centering\arraybackslash}p{2.2cm} | %
        cccc@{}}
        \hline
        $f$ [Hz] & Metrics & PID & NMPC & MI-NMPC & DM-SMC \\
        \hline
        \multirow{4}{*}{0.10}
        & AE [kPa]           & 1.64   & 1.14   & 0.78   & \textbf{0.72}   \\
        & $\max |e|$ [kPa]   & 7.19   & 13.48  & \textbf{3.89}  & 5.09   \\
        & PWM-E [\%$\cdot$s] & 268.80 & 258.15 & 256.11 & \textbf{248.89} \\
        & Switches           & 14.67  & 10.00  & 37.00  & \textbf{2.00}   \\
        \hline
        \multirow{4}{*}{0.25}
        & AE [kPa]           & 1.61   & 1.33   & 1.20   & \textbf{0.78}   \\
        & $\max |e|$ [kPa]   & 6.72   & 6.52   & \textbf{4.71}  & 5.79   \\
        & PWM-E [\%$\cdot$s] & 119.07 & 113.09 & 111.71 & \textbf{108.81} \\
        & Switches           & 7.00   & 4.67   & 16.67  & \textbf{2.00}   \\
        \hline
        \multirow{4}{*}{0.50}
        & AE [kPa]           & 1.80   & 2.17   & 2.18   & \textbf{0.86}   \\
        & $\max |e|$ [kPa]   & 7.68   & 7.31   & \textbf{5.52}  & 6.66   \\
        & PWM-E [\%$\cdot$s] & 70.25  & 66.31  & \textbf{65.35} & 65.59  \\
        & Switches           & 3.33   & 2.67   & 9.33   & \textbf{2.00}   \\
        \hline
    \end{tabular}%
    }
    \vspace{-4mm}
\end{table}

\section{Experimental Results}
\subsection{Experimental Setup}
\begin{figure}[t]
  \centering
  \includegraphics[width=0.98\columnwidth]{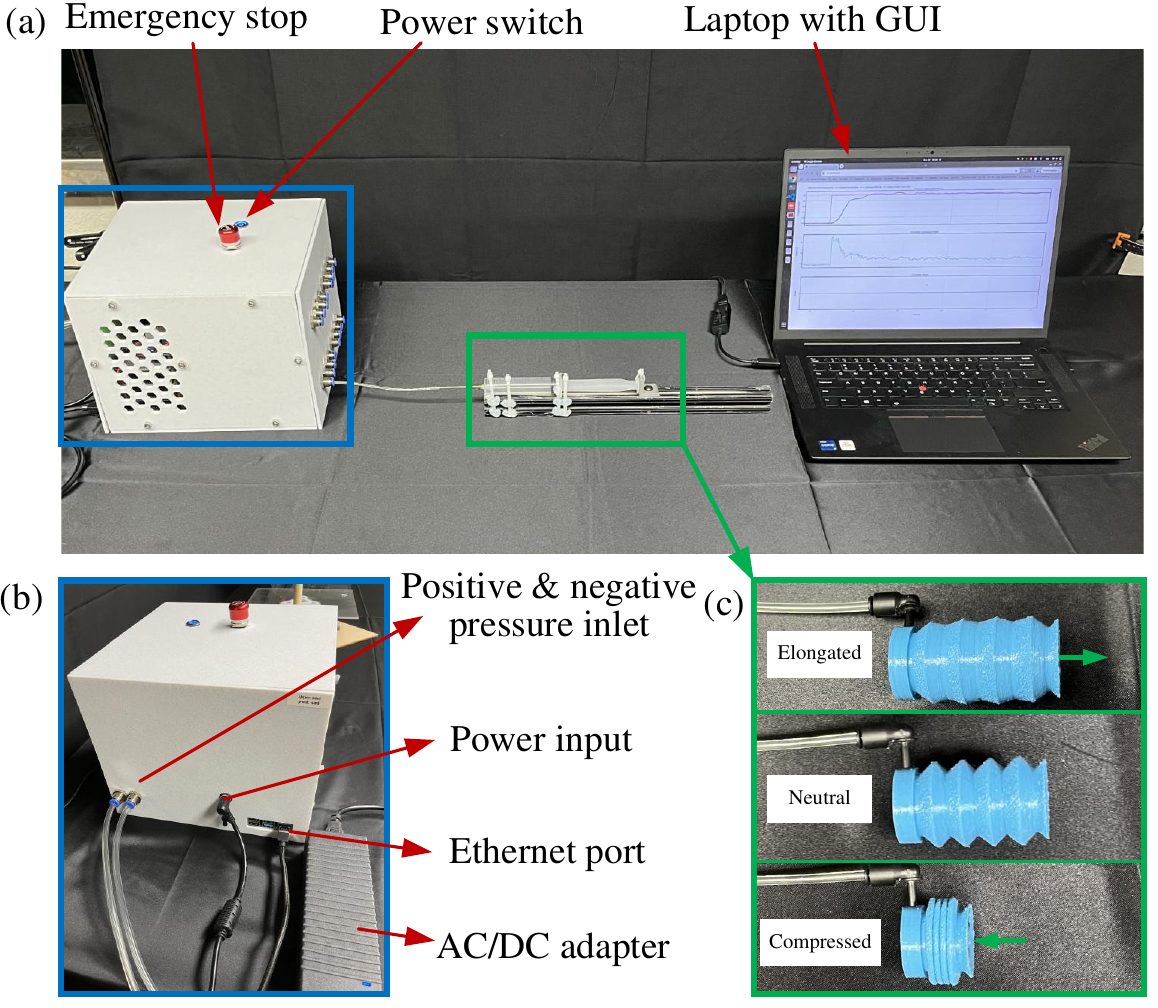}
  \caption{Experimental setup for BiPneu and DM-SMC evaluation. (a) BiPneu with fixed-volume load (20~mL syringe). (b) Rear view. (c) Soft bellow actuator as a varying-volume load.}
  \label{fig:setup}
  \vspace{-3mm}
\end{figure}

To validate the proposed BiPneu system and dual-mode SMC, we built the experimental platform shown in Fig.~\ref{fig:setup}. The BiPneu prototype is shown in Fig.~\ref{fig:setup}(a)--(b) and an emergency stop is designed to ensure safe outflow, which is particularly important for rehabilitation applications like soft exoskeleton gloves \cite{hu2020novel}. The pressure feedback is sampled at 60~Hz, limited by the integrated ADC, while control commands are updated at 200~Hz. Fig.~\ref{fig:setup}(b) presents the rear view of BiPneu, highlighting the pneumatic and electrical interfaces. As shown in Fig.~\ref{fig:setup}(a), a 20~mL syringe with its piston rigidly fixed is connected to Channel~1 of BiPneu, providing a constant-volume load. This configuration is used to evaluate step and sinusoidal pressure-tracking performance. For the switched model in Eq.~\eqref{eq:controlform}, the outlet volume is set to \num{2.0e-5}~\si{m^{3}} (20~mL), which serves as a representative test case. The same modeling and control framework can be applied to other load volumes once estimated.

The sonic conductances $C_{po}$, $C_{on}$, $C_{oa}$, and $C_{ao}$ in Eq.~\eqref{eq:Afunctions} are identified offline from step-response pressure data via least-squares fitting. The valve spool-fraction mapping $\bar{x}(u)$ is calibrated separately for inflation and deflation using a cubic polynomial regression, $\bar{x}(u)=\sum_{k=0}^{3} a_ku^k$, based on step tests at different PWM duty cycles $u$. The identified parameters are summarized in Table~\ref{tab:sys_id_params_full}, and further identification details are provided in the Section~A of Supplementary Material.

\begin{table}[t]
\caption{Identified sonic conductances and calibrated spool-fraction mapping coefficients.}
\label{tab:sys_id_params_full}
\centering
\renewcommand{\arraystretch}{1.0}
\begin{tabular}{p{2.5cm} p{5.2cm}}   
\toprule
\textbf{Parameter} & \textbf{Value} \\
\midrule
Sonic conductances
& $\begin{aligned}
C_{po} &= 2.64\times10^{-10}\ \mathrm{m^3\,s^{-1}\,Pa^{-1}}\\
C_{on} &= 3.44\times10^{-10}\ \mathrm{m^3\,s^{-1}\,Pa^{-1}}\\
C_{oa} &= 6.94\times10^{-12}\ \mathrm{m^3\,s^{-1}\,Pa^{-1}}\\
C_{ao} &= 4.52\times10^{-12}\ \mathrm{m^3\,s^{-1}\,Pa^{-1}}
\end{aligned}$ \\
\midrule
\parbox[t]{3.5cm}{Spool-fraction\\mapping coefficients}
&
\begin{tabular}[t]{@{}l l@{}}
\textbf{Inflation} & \textbf{Deflation} \\
$a_3 = 4.38\times10^{-6}$ & $a_3 = 6.95\times10^{-6}$ \\
$a_2 = -1.09\times10^{-3}$ & $a_2 = -1.62\times10^{-3}$ \\
$a_1 = 8.96\times10^{-2}$ & $a_1 = 1.25\times10^{-1}$ \\
$a_0 = -1.48$ & $a_0 = -2.27$
\end{tabular}
\\
\bottomrule
\end{tabular}
\vspace{-5mm}
\end{table}

\subsection{Pressure Tracking under Fixed-Volume Load}
\label{sec:pressure_tracking}
As discussed in the simulation studies, both NMPC and MI-NMPC are difficult to run in real time on the embedded system, so we implement and compare DM-SMC and PID only on the BiPneu platform. Experiments are first conducted using the setup with constant volume load as seen in Fig.~\ref{fig:setup}(a). The same multi-step and sinusoidal reference trajectories as in simulation are used. Both controllers employ the same hysteresis-supervised mode switching shown in Eq.~\eqref{eq:mode_selection}, and thus use well-tuned mode-dependent parameters for inflation ($I$) and deflation ($D$). For PID, the gains are $(K_p^{(I)},K_i^{(I)},K_d^{(I)};\;K_p^{(D)},K_i^{(D)},K_d^{(D)}) = (0.32,0.3,0.02;\;0.6,0.2,0.01)$. For DM-SMC, the inflation-mode parameters are $\lambda^{(I)}=2.8$, $\eta^{(I)}=5.0{\times}10^3$, $\mu^{(I)}=1.0{\times}10^3$, and $k_I^{(I)}=0.8$, while the deflation-mode parameters are $\lambda^{(D)}=4.0$, $\eta^{(D)}=5.0{\times}10^3$, $\mu^{(D)}=1.0{\times}10^3$, and $k_I^{(D)}=0.8$. The controller parameters, including the PID gains and DM-SMC weights, were tuned empirically to achieve good overall trade-off among transient response, steady-state error, and overshoot. The hysteresis half-band is set to $h=5$~kPa, selected empirically to balance switching robustness and tracking performance.

\begin{figure}[!htbp]
\centerline{\includegraphics[scale=0.41]{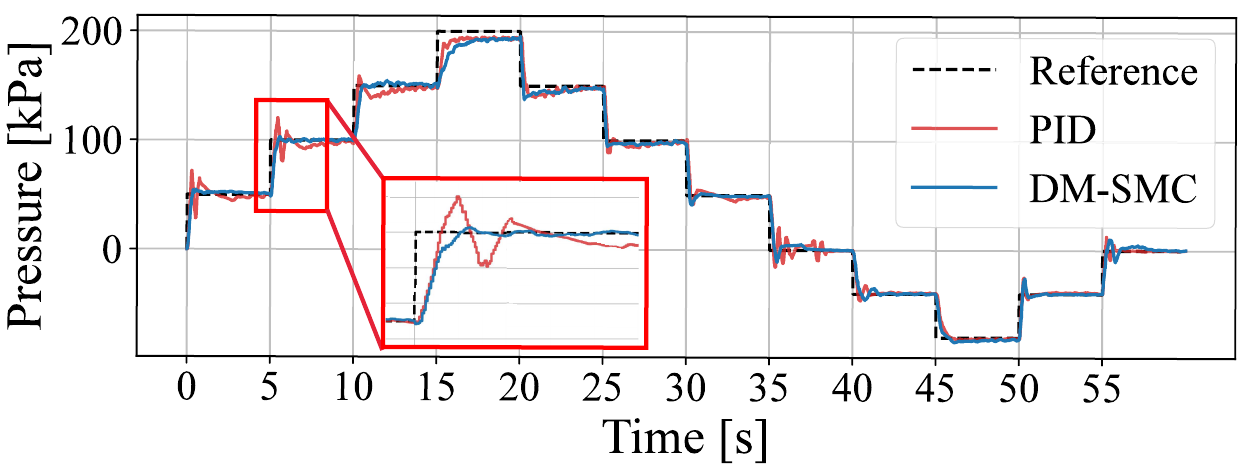}}
\vspace{-2mm}
\caption{Experimental results of multi-step reference-tracking for PID and DM-SMC.}
\label{fig:Exp_Step_DM}
\vspace{-6mm}
\end{figure}

\begin{figure}[!htbp]
\centerline{\includegraphics[scale=0.315]{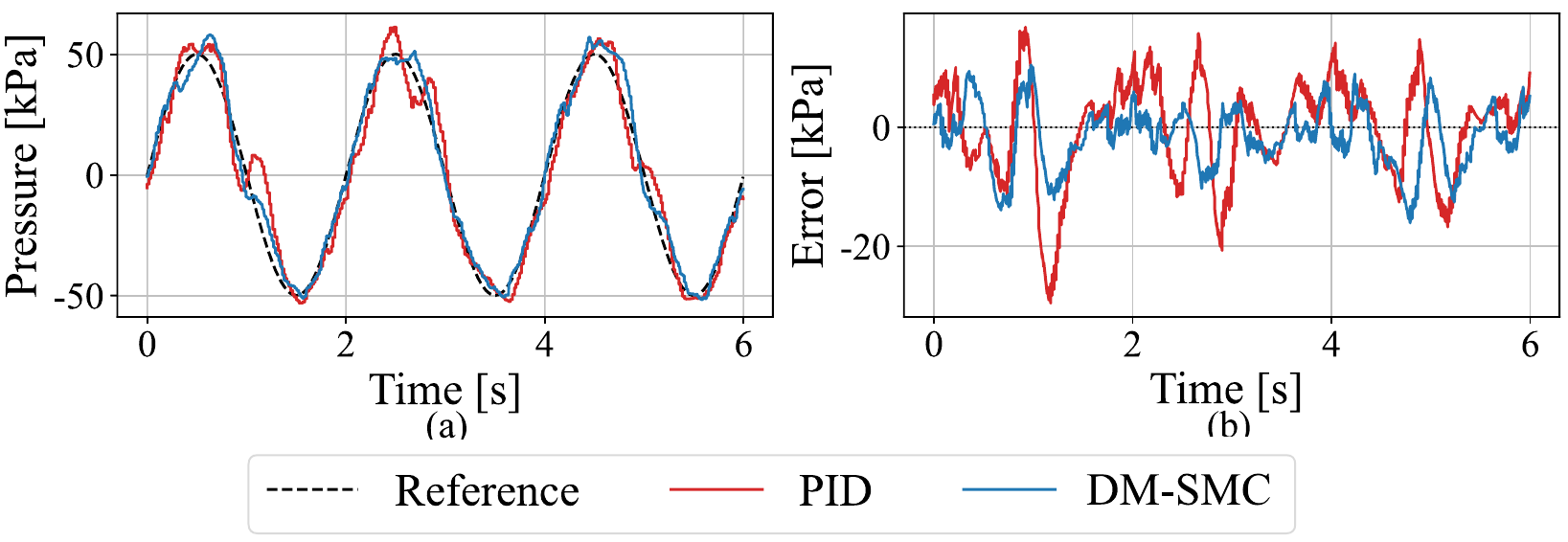}}
\vspace{-2mm}
\caption{Experimental results of sinusoidal reference-tracking at 0.5 Hz for PID and DM-SMC: (a) pressure trajectories; (b) corresponding tracking errors.}
\label{fig:Exp_Sine_DM}
\vspace{-3mm}
\end{figure}

Experimental results are shown in Fig.~\ref{fig:Exp_Step_DM} and Fig.~\ref{fig:Exp_Sine_DM}, and the corresponding quantitative comparisons are summarized in Table~\ref{tab:Exp_Step_Merged} and Table~\ref{tab:Exp_Sine_Merged}, under the columns ``Fixed-volume". For brevity, Fig.~\ref{fig:Exp_Sine_DM} shows only the representative 0.5~Hz case and other sinusoidal tracking responses are reported in Fig.~\ref{fig:supp_Exp_Sine}. The same performance metrics are reported as in simulation. In the multi-step experiments, DM-SMC exhibits consistently better transient and steady tracking behavior than PID, as shown in Table~\ref{tab:Exp_Step_Merged} and Fig.~\ref{fig:Exp_Step_DM}. Compared with PID, DM-SMC reduces AE by 11.9\%, and decreases ITAE by 51.4\%, indicating faster attenuation of tracking error during transients. Meanwhile, DM-SMC slightly reduces control effort (PWM-E) by 4.5\% and requires 51.4\% fewer mode switches. In the sinusoidal tracking experiments (Table~\ref{tab:Exp_Sine_Merged} and Fig.~\ref{fig:Exp_Sine_DM}), DM-SMC achieves comparable performance to PID at $0.1$~Hz in terms of AE, while the differences become increasingly evident as the frequency increases. At $0.25$~Hz and $0.5$~Hz, DM-SMC reduces AE by 32.19\% and 35.62\%, respectively. It also reduces max$|e|$ by 6.37\% at $0.25$~Hz and by 45.50\% at $0.5$~Hz, while also lowering PWM-E and reducing switching. Notably, DM-SMC achieves average absolute errors of 1.44~kPa and 4.23~kPa in the multi-step and sinusoidal tests, corresponding to 0.5\% and 1.5\% of the full pressure range. The relatively larger error at $0.1$~Hz is likely due to the higher mismatch of the identified mapping $\bar{x}(u)$ under slowly varying references, which would lead to the observed non-monotonic trend across frequencies. Besides, the discrepancy between simulations and experiments can be attributed to system identification errors and practical uncertainties, such as sensor noise and tubing compliance. This discrepancy is more evident in ITAE because the metric weighs the error by time.

A plausible explanation for DM-SMC outperforming PID is its stronger correction and higher robustness to modeling uncertainty and disturbances, together with explicit use of the identified pressure dynamics as feedforward compensation.

\begin{table}[!htbp]
    \centering
    \caption{Experimental multi-step reference tracking performance under fixed-volume and varying-volume loads: PID vs.\ DM-SMC.}
    \label{tab:Exp_Step_Merged}
    \renewcommand{\arraystretch}{1.2}%
    \resizebox{0.95\columnwidth}{!}{%
    \begin{tabular}{@{}>{\centering\arraybackslash}p{2.8cm}|cc|cc@{}}
        \hline
        \multirow{2}{*}{Metrics}
        & \multicolumn{2}{c|}{Fixed-volume}
        & \multicolumn{2}{c@{}}{Varying-volume} \\
        \cline{2-5}
        & PID & DM-SMC & PID & DM-SMC \\
        \hline
        $e_{\mathrm{ss}}$ [kPa]         & 1.52  & \textbf{1.44} & 1.62  & \textbf{1.43} \\
        AE [kPa]                        & 2.87  & \textbf{2.53} & 2.72  & \textbf{2.59} \\
        ITAE [kPa$\cdot$s$^2$]          & 57.99 & \textbf{28.18} & \textbf{26.41} & 27.27 \\
        PWM-E [\%$\cdot$s]              & 132.93 & \textbf{126.89} & 134.30 & \textbf{129.01} \\
        Switches                        & 2.92  & \textbf{1.42} & 2.75  & \textbf{1.33} \\
        \hline
    \end{tabular}%
    }
\end{table}

\begin{table}[!htbp]
    \centering
    \caption{Experimental sinusoidal reference tracking performance under fixed-volume and varying-volume loads: PID vs.\ DM-SMC.}
    \label{tab:Exp_Sine_Merged}
    \renewcommand{\arraystretch}{1.2}%
    \resizebox{0.95\columnwidth}{!}{%
    \begin{tabular}{@{}>{\centering\arraybackslash}p{0.9cm}>{\centering\arraybackslash}p{2.0cm}|cc|cc@{}}
        \hline
        \multicolumn{1}{c}{\multirow{2}{*}{$f$ [Hz]}} 
        & \multicolumn{1}{c|}{\multirow{2}{*}{Metrics}}
        & \multicolumn{2}{c|}{Fixed-volume}
        & \multicolumn{2}{c@{}}{Varying-volume} \\
        \cline{3-6}
        & & PID & DM-SMC & PID & DM-SMC \\
        \hline
        \multirow{4}{*}{0.10}
        & AE [kPa]               & \textbf{4.00} & 4.01 & \textbf{3.62} & 4.19 \\
        & $\max |e|$ [kPa]       & \textbf{13.88} & 18.73 & \textbf{14.15} & 15.21 \\
        & PWM-E [\%$\cdot$s]     & 250.07 & \textbf{247.39} & \textbf{251.09} & 256.23 \\
        & Switches               & \textbf{4.00} & 4.67 & 3.33 & \textbf{2.00} \\
        \hline
        \multirow{4}{*}{0.25}
        & AE [kPa]               & 4.38 & \textbf{2.97} & 3.60 & \textbf{2.16} \\
        & $\max |e|$ [kPa]       & 16.80 & \textbf{15.73} & 19.49 & \textbf{8.92} \\
        & PWM-E [\%$\cdot$s]     & 107.07 & \textbf{104.11} & 108.55 & \textbf{107.41} \\
        & Switches               & 2.67 & \textbf{2.00} & 2.67 & \textbf{2.00} \\
        \hline
        \multirow{4}{*}{0.50}
        & AE [kPa]               & 6.57 & \textbf{4.23} & 7.17 & \textbf{3.60} \\
        & $\max |e|$ [kPa]       & 29.54 & \textbf{16.10} & 40.43 & \textbf{14.37} \\
        & PWM-E [\%$\cdot$s]     & 62.16 & \textbf{57.97} & 65.16 & \textbf{60.45} \\
        & Switches               & 3.33 & \textbf{2.00} & 2.00 & 2.00 \\
        \hline
    \end{tabular}%
    }
    \vspace{-5mm}
\end{table}

\subsection{Robustness to Varying-Volume Load}
\label{sec:robustness}
To further evaluate the robustness of the proposed DM-SMC to load variations, we connect BiPneu to a 3D-printed soft bellow actuator fabricated from TPU 82A (Filaflex 82A), which introduces a deformation-dependent outlet volume. Fig.~\ref{fig:setup}(c) shows three representative volume states of the bellow actuator, including compressed, neutral, and elongated, with its internal volume varying from approximately $0$ to $25$~mL. This varying-volume load enables evaluation of the system and controller on a practical soft-robot actuator whose internal volume changes during motion. The reference trajectories and all controller parameters are the same as those used with the fixed 20~mL load in Sec.~\ref{sec:pressure_tracking}. Quantitative results are summarized in Tables~\ref{tab:Exp_Step_Merged} and~\ref{tab:Exp_Sine_Merged} under the ``Varying-volume'' columns, while representative plots are provided in the supplementary material (see Figs.~\ref{fig:supp_Exp_Step_Bellow} and~\ref{fig:supp_Exp_Sine_Bellow}).

Across both multi-step and sinusoidal tests, DM-SMC outperforms PID under the deformation-induced varying-volume load, demonstrating robustness to load-dependent changes in pneumatic dynamics. This robustness is mainly attributed to the mode-specific sliding-mode design, which maintains corrective action against modeling mismatch within each mode, together with the hysteresis-supervised mode switching that avoids unnecessary inflation/deflation toggling under varying load conditions. In particular, DM-SMC maintains superior transient responses during step tracking with fewer mode switches, and its advantage in sinusoidal tracking becomes more pronounced at higher frequencies, where PID exhibits larger phase lag and peak errors. In several cases, DM-SMC under the varying-volume bellow load achieves performance comparable to, or slightly better than, fixed-volume results, possibly due to added compliance and damping from bellow deformation that smooth the pressure response.

\section{Demonstration of BiPneu in Soft Robot Applications}
\subsection{Ball Maneuvering with a Soft Parallel Manipulator}
\begin{figure}[!t]
  \centering
  \includegraphics[width=0.96\columnwidth]{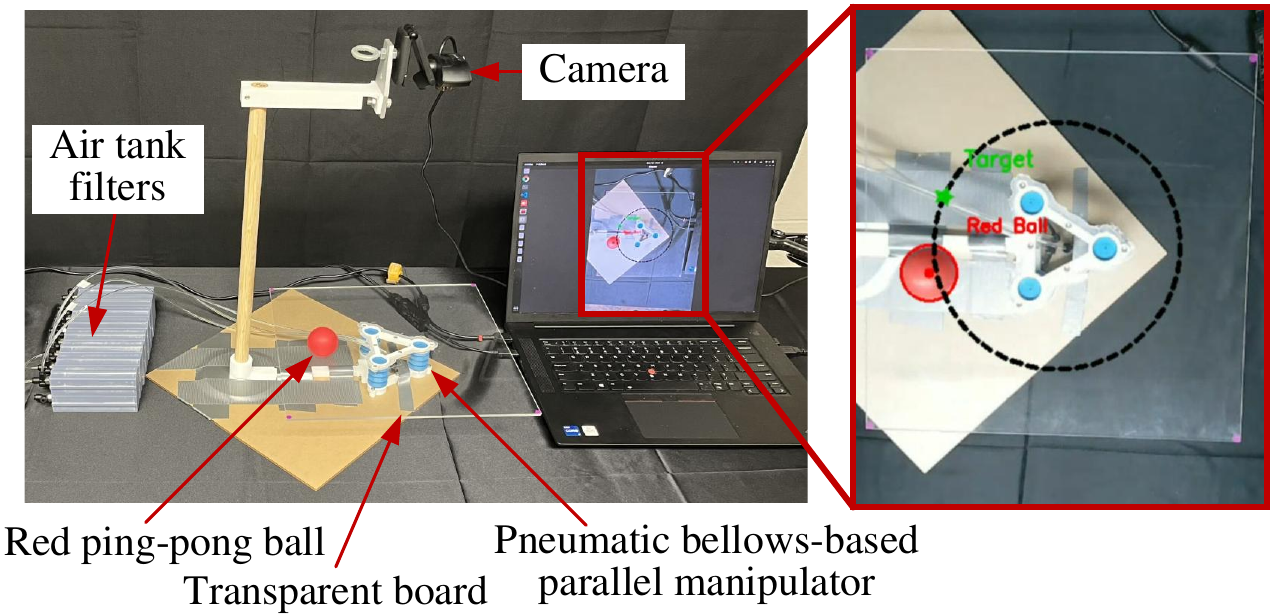}
  \vspace{-1mm}
  \caption{Setup for ball maneuvering with a parallel manipulator.}
  \label{fig:application_ball}
  \vspace{-6mm}
\end{figure}
Ball-and-plate systems are typically realized using traditional rigid motors in prior studies, because this is a highly dynamic task requiring fast and accurate tilt regulation. Here, we realize this task using a soft parallel manipulator driven by BiPneu. As shown in Fig.~\ref{fig:application_ball}, a red ping-pong ball is maneuvered on a transparent board supported by three parallel bellow actuators to reach target locations and track fast reference trajectories. Differential pressurization tilts the board and drives the ball motion. An overhead camera estimates the ball position at 60~Hz, and a PID outer-loop computes the desired board tilt. A piecewise-constant-curvature (PCC)–based inverse-kinematics optimization scheme then determines desired bellow lengths, which are converted to pressure references via a calibrated mapping and sent to BiPneu. BiPneu independently regulates the three chambers using DM-SMC. 

\begin{figure}[!t]
  \centering
  \includegraphics[width=0.9\columnwidth]{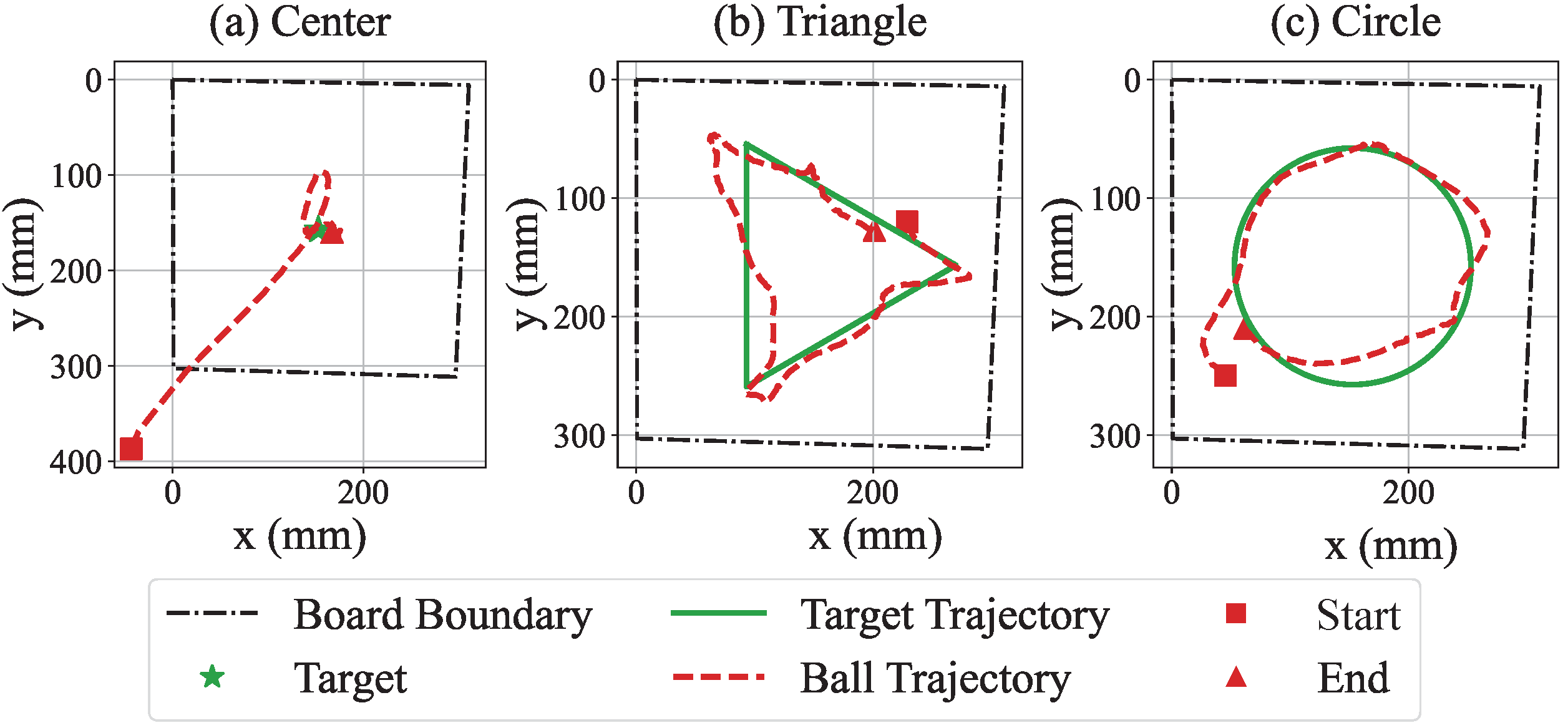}
  \vspace{-2mm}
  \caption{Results of ball maneuvering in different tracking tasks.}
  \label{fig:ballbalance}
  \vspace{-4mm}
\end{figure}

The supplementary video\footnote{[Online]. Available: \href{https://youtu.be/ujWjucfVKs0}{https://youtu.be/ujWjucfVKs0}} further demonstrates BiPneu with DM-SMC in ball-manipulation tasks, including target regulation, disturbance rejection and fast trajectory tracking. Fig.~\ref{fig:ballbalance} shows representative trials. In center regulation, the ball converges to the target with a final error of 1.4~cm (3.3\% of the 43~cm board diagonal). For triangular and circular tracking, root-mean-square errors (RMSE) of 3.0~cm (7.0\%) and 3.3~cm (7.7\%) are achieved, respectively. These results demonstrate that BiPneu can achieve accurate and responsive pressure regulation in the challenging dynamic interaction task.

\vspace{-3mm}
\subsection{Teleoperation of a Soft Bellow Actuator via Real-time FEM Simulator}
\begin{figure}[!t]
  \centering
  \includegraphics[width=0.88\columnwidth]{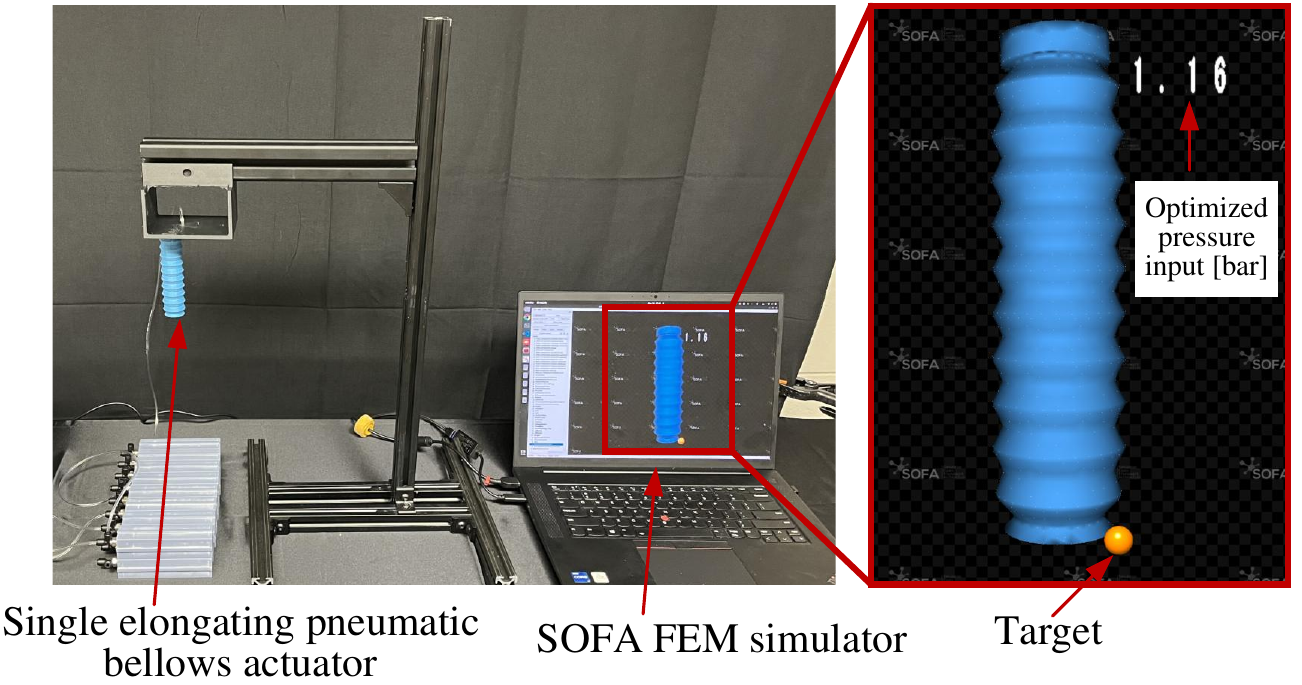}
  \vspace{-1mm}
  \caption{Demonstration setup of real-time FEM-based teleoperation of a soft bellows actuator.}
  \label{fig:application_teleoperation}
  \vspace{-6mm}
\end{figure}

We demonstrate BiPneu’s software ecosystem compatibility through real-time FEM-based teleoperation of a soft bellow actuator, which enables efficient data collection for training control policies but is challenging by inverse-kinematics complexity and sensitivity to actuation latency. Leveraging its accurate and fast pressure regulation and seamless integration with a high-fidelity physical simulator, we achieve responsive teleoperation with minimal latency, as shown in Fig.~\ref{fig:application_teleoperation}.

The actuator is modeled in the FEM simulator SOFA \cite{faure2012sofa}, where desired actuator lengths are specified interactively and converted to pressure commands via inverse optimization in SOFA, then streamed to BiPneu through ROS~2 for closed-loop control. Across 32 target lengths spanning $-75$ to $200$~kPa, the actuator tracks the SOFA reference with an RMSE of 4.1~mm (5.56\% of the 73~mm workspace), as detailed in Fig.~\ref{fig:supp_SOFA_results} in the Supplementary Material. The supplementary video further confirms the low end-to-end latency, enabling effective real-time teleoperation.

\vspace{-4mm}
\section{Conclusion}
In this work, we developed BiPneu as a practical and extensible platform for multi-channel bipolar-pressure actuation in soft robotic systems. The proposed hybrid electro-pneumatic modeling and dual-mode SMC framework achieves accurate and responsive pressure regulation under both fixed- and varying-volume conditions, addressing a key challenge in real-world soft actuators. Experimental results further demonstrate its effectiveness, enabling dynamic interaction tasks and low-latency FEM-based teleoperation.

For future work, we will leverage BiPneu for advanced control applications of soft robots, such as developing VLA policies for soft robots and exploring learning-based control with efficient sim-to-real policy transfer.

\vspace{-5mm}
\bibliographystyle{IEEEtran}
\bibliography{TMech}

\clearpage
\setcounter{page}{1}

\section*{Supplementary Material}
\label{sec:supp_sys_id}

\setcounter{figure}{0}
\setcounter{table}{0}
\setcounter{equation}{0}
\renewcommand{\thefigure}{S\arabic{figure}}
\renewcommand{\thetable}{S\arabic{table}}
\renewcommand{\theequation}{S\arabic{equation}}

\subsection{System Identification}
\begin{figure}[b]
  \centering
  \includegraphics[width=0.75\linewidth]{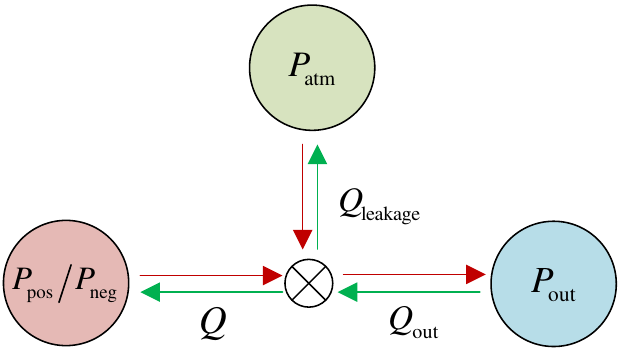}
  \caption{Schematic of the mass-flow contributions to the air receiver.}
  \label{figS:flowgraph}
\end{figure}

This subsection describes how we identify the sonic conductances $C_{po}$, $C_{on}$, $C_{oa}$, and $C_{ao}$, and how we separately calibrate the spool-fraction mapping $\bar{x}(u)$ for inflation and deflation. As we studies in the previous work \cite{mei2025modeling}, the effective mass flow rate $Q_{\text{out}}$ delivered to the air receiver is determined by two contributions: the bidirectional mass flow $Q$ through Valve~1, which accounts for both inflation and deflation mode, and the leakage flow $Q_{\text{leakage}}$ through Valve~2 to the atmosphere, as illustrated in Fig.~\ref{figS:flowgraph}.

\begin{figure}[t]
  \centering
  \includegraphics[width=0.95\linewidth]{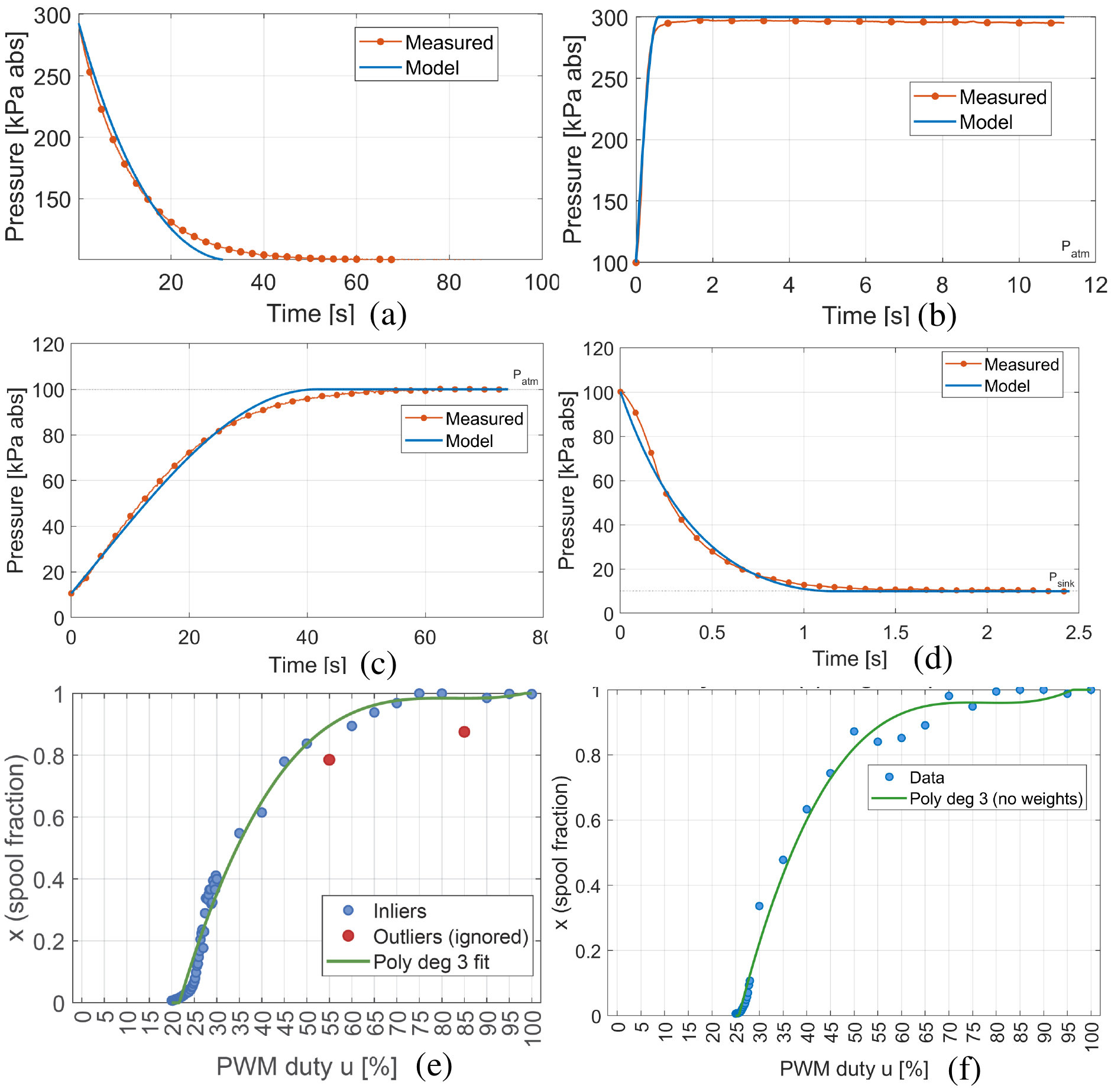}
  \caption{Calibration and identification results. (a) Leakage-only decay ($\bar{x}=0$) for the inflation case (Case~1), used to identify the outlet-to-atmosphere sonic conductance $C_{oa}$. (b) Inflow-only rise ($\bar{x}=1$) for the inflation case, used to identify the supply-to-outlet sonic conductance $C_{po}$. (c) Leakage-only decay ($\bar{x}=0$) for the deflation case (Case~4), used to identify the outlet-to-atmosphere sonic conductance $C_{oa}$. (d) Outflow-only decay ($\bar{x}=1$) for the deflation case, used to identify the outlet-to-vacuum sonic conductance $C_{on}$. (e) Segment-wise identified spool fractions $\hat{x}_i$ versus PWM duty $u_i$ for inflation and the resulting cubic fit $\bar{x}(u)$. (f) Segment-wise identified spool fractions $\hat{x}_i$ versus PWM duty $u_i$ for deflation and the resulting cubic fit $\bar{x}(u)$.}
  \label{figS:calibration}
\end{figure}

Depending on the receiver pressure $P_{\text{out}}$ relative to the positive supply $P_{\text{pos}}$, negative sink $P_{\text{neg}}$, and atmosphere $P_{\text{atm}}$, there exist four different operating cases: 

\begin{subequations}\label{eq:qout_cases}
\begin{enumerate}

  \item Inflation process with $P_{\text{atm}} < P_{\text{out}} < P_{\text{pos}}$:
  \begin{equation}
    \resizebox{0.9\columnwidth}{!}{$
    \begin{aligned}
      Q_{\text{out}} &= Q - Q_{\text{leakage}} \\
                       &= Q(\bar{x};\, P_{\text{pos}},\, P_{\text{out}},\, C_{po}) 
                        \;-\; Q(1-\bar{x};\, P_{\text{out}},\, P_{\text{atm}},\, C_{oa}),
    \end{aligned}
    $}
  \end{equation}

  \item Inflation process with $P_{\text{out}} < P_{\text{atm}} < P_{\text{pos}}$:
  \begin{equation}
    \resizebox{0.9\columnwidth}{!}{$
    \begin{aligned}
      Q_{\text{out}} &= Q + Q_{\text{leakage}} \\
                       &= Q(\bar{x};\, P_{\text{pos}},\, P_{\text{out}},\, C_{po})
                        \;+\; Q(1-\bar{x};\, P_{\text{atm}},\, P_{\text{out}},\, C_{ao}),
    \end{aligned}
    $}
  \end{equation}

  \item Deflation process with $P_{\text{neg}} < P_{\text{atm}} < P_{\text{out}}$:
  \begin{equation}
    \resizebox{0.9\columnwidth}{!}{$
    \begin{aligned}
      Q_{\text{out}} &= -\,Q - Q_{\text{leakage}} \\
                       &= -\,Q(\bar{x};\, P_{\text{out}},\, P_{\text{neg}},\, C_{on})
                         \;-\; Q(1-\bar{x};\, P_{\text{out}},\, P_{\text{atm}},\, C_{oa}),
    \end{aligned}
    $}
  \end{equation}

  \item Deflation process with $P_{\text{neg}} < P_{\text{out}} < P_{\text{atm}}$:
  \begin{equation}
    \resizebox{0.9\columnwidth}{!}{$
    \begin{aligned}
      Q_{\text{out}} &= -\,Q + Q_{\text{leakage}} \\
                       &= -\,Q(\bar{x};\, P_{\text{out}},\, P_{\text{neg}},\, C_{on})
                         \;+\; Q(1-\bar{x};\, P_{\text{atm}},\, P_{\text{out}},\, C_{ao}),
    \end{aligned}
    $}
  \end{equation}
\end{enumerate}
\end{subequations}
In this study, we use Cases 1 and 4 to identify the four sonic conductances $C_{po}$, $C_{on}$, $C_{oa}$, and $C_{ao}$. These two cases are easy to realize experimentally and allow reliable, repeatable step-response data collection. And the spool-fraction mapping is represented by a cubic polynomial: \begin{equation} \bar x(u)=\sum_{k=0}^{3} a_k\,u^k, \end{equation} whose coefficients $\{a_k\}$ are calibrated separately for inflation and deflation from step-response data.

In Case~1, we identify $C_{oa}$, $C_{po}$, and the spool fraction $\bar{x}$ during the inflation process using a three-step procedure. The Case~1 dataset is collected by recording step responses with $u_1 = 100\%$ while sweeping $u_2$ from $20\%$ to $30\%$ in $0.2\%$ increments and from $30\%$ to $100\%$ in $5\%$ increments. For each step response, once the pressure reaches steady state, $u_2$ is set to zero to allow the pressure to decay through leakage, and the resulting decay trajectory is recorded.

First, we isolate the \emph{leakage branch} by setting $\bar{x}=0$, which closes the supply path and leaves only the outlet-to-atmosphere discharge through Valve~2. Therefore, the dynamics of outlet pressure turns out to be: 
\begin{equation}\label{eq:leakage_dyn}
\left\{
\begin{aligned}
Q_{\text{out}} \big|_{\bar{x}=0}
&= Q(0;\, P_{\text{pos}},\, P_{\text{out}},\, C_{po})
   - Q(1;\, P_{\text{out}},\, P_{\text{atm}},\, C_{oa}) \\
&= -\,Q(1;\, P_{\text{out}},\, P_{\text{atm}},\, C_{oa}), \\[4pt]
\dot{P}_{\text{out}}
&= \frac{\gamma R T}{V}\, Q_{\text{out}}.
\end{aligned}
\right.
\end{equation}
Based on the measured decay process, the outlet conductance $C_{oa}$ is identified by solving a
nonlinear least-squares problem that minimizes the mismatch between the measured outlet pressure and
the model-predicted pressure obtained from \eqref{eq:leakage_dyn}. Specifically, given the sampled
decay trajectory $\{(t_i, P_{\text{out}}(t_i))\}_{i=1}^{N}$, we estimate $C_{oa}$ as
\begin{equation}\label{eq:ls_Coa}
\hat C_{oa}
= \arg\min_{C_{oa}>0}
\sum_{i=1}^{N}
\Big(
\hat P_{\text{out}}(t_i; C_{oa}) - P_{\text{out}}(t_i)
\Big)^2,
\end{equation}
where $\hat P_{\text{out}}(t_i; C_{oa})$ denotes the simulated outlet pressure at time $t_i$, obtained
by numerically integrating \eqref{eq:leakage_dyn} with the initial condition
$P_{\text{out}}(0)=P_{\text{out}}(t_1)$. The fitting result is shown in Fig.~\ref{figS:calibration} (a), and the numerical result is listed in Tab.~\ref{tab:sys_id_params_full}.

Next, we isolate the \emph{inflow branch} by setting $\bar{x}=1$, which closes the leakage branch and produces pure inflation from $P_{\text{pos}}$ to the outlet. As a result, the outlet load undergoes pure inflation from the positive-pressure source $P_{\text{pos}}$, and the outlet flow reduces to:
\begin{equation}\label{eq:inflow_dyn}
\left\{
\begin{aligned}
Q_{\text{out}} \big|_{\bar{x}=1}
&= Q(1;\, P_{\text{pos}},\, P_{\text{out}},\, C_{po})
   - Q(0;\, P_{\text{out}},\, P_{\text{atm}},\, C_{oa}) \\
&= \,Q(1;\, P_{\text{pos}},\, P_{\text{out}},\, C_{po}), \\[4pt]
\dot{P}_{\text{out}}
&= \frac{\gamma R T}{V}\, Q_{\text{out}}.
\end{aligned}
\right.
\end{equation}
Based on the measured pressure-rise process, the supply-to-outlet conductance $C_{po}$ is identified
by solving a nonlinear least-squares problem that minimizes the mismatch between the measured outlet
pressure and the model-predicted pressure obtained from \eqref{eq:inflow_dyn}. Specifically, given
the sampled rise trajectory $\{(t_i, P_{\text{out}}(t_i))\}_{i=1}^{N}$, we estimate $C_{po}$ as
\begin{equation}\label{eq:ls_Cpo}
\hat C_{po}
= \arg\min_{C_{po}>0}
\sum_{i=1}^{N}
\Big(
\hat P_{\text{out}}(t_i; C_{po}) - P_{\text{out}}(t_i)
\Big)^2,
\end{equation}
where $\hat P_{\text{out}}(t_i; C_{po})$ denotes the simulated outlet pressure at time $t_i$, obtained
by numerically integrating \eqref{eq:inflow_dyn} with the initial condition
$P_{\text{out}}(0)=P_{\text{out}}(t_1)$. The fitting result is shown in Fig.~\ref{figS:calibration} (b), and the numerical result is listed in Tab.~\ref{tab:sys_id_params_full}.

Finally, with $C_{oa}$ and $C_{po}$ fixed, we identify the spool fraction $\bar{x}$ in each constant-PWM segment by matching the measured pressure evolution to the pressure-dynamics model
\begin{equation}
\dot{P}_{\mathrm{out}}
=\gamma\frac{RT}{V}\Big(Q\!\left(t;\bar{x}(u)\right)-Q_{\mathrm{leakage}}\!\left(t;1-\bar{x}(u)\right)\Big),
\end{equation}
where $Q(\cdot)$ and $Q_{\mathrm{leakage}}(\cdot)$ denote the inlet and leakage mass-flow terms, respectively, parameterized by the effective port openings $\bar{x}(u)$ and $1-\bar{x}(u)$.
For the $i$-th PWM segment with samples $\{t_k\}$, we estimate $\hat{x}_i$ by the segment-wise least-squares problem
\begin{equation}
\hat{x}_i
=\arg\min_{0<x<1}\sum_{k}\Big(P_{\mathrm{out}}(t_k;x)-P_{\mathrm{meas}}(t_k)\Big)^2.
\end{equation}
This yields a set of calibration pairs $\{(u_i,\hat{x}_i)\}$, which we fit with a cubic polynomial to obtain the inflation mapping $\bar{x}(u)$. The fitting result is shown in Fig.~\ref{figS:calibration} (e), and the numerical result is listed in Tab.~\ref{tab:sys_id_params_full}.

A similar three-step identification procedure is used for the deflation case (Case~4) to identify $C_{ao}$, $C_{on}$, and the corresponding spool-fraction mapping $\bar{x}(u)$. Specifically, we collect step-response data by setting $u_1=0\%$ and sweeping $u_2$ across multiple duty levels, so that the outlet is regulated by the negative-pressure branch while Valve~2 still contributes atmospheric exchange. We then (i) isolate the atmosphere-to-outlet branch to estimate $C_{ao}$ (Fig.~\ref{figS:calibration}(c)), (ii) isolate the outlet-to-sink branch to estimate $C_{on}$ (Fig.~\ref{figS:calibration}(d)), and (iii) with $C_{ao}$ and $C_{on}$ fixed, perform segment-wise least-squares identification of $\bar{x}$ for each constant-PWM segment and fit the resulting ${(u_i,\hat{x}_i)}$ pairs with a cubic polynomial to obtain the deflation mapping $\bar{x}(u)$ (Fig.~\ref{figS:calibration}(f)). All the numerical results are listed in Tab.~\ref{tab:sys_id_params_full}. 

\subsection{ROS2 Software Architecture and Message Interface}
\label{sec:supp_software}

This section details the ROS2 node organization and message-level communication used in the distributed BiPneu software architecture shown in Fig.~\ref{fig:Software}. The architecture follows a server--client paradigm, where time-critical sensing, control, and actuation are executed on the embedded client, while high-level command generation, visualization, and data logging are handled on the server.

\begin{figure}[h]
    \centering
    \includegraphics[width=1.0\columnwidth]{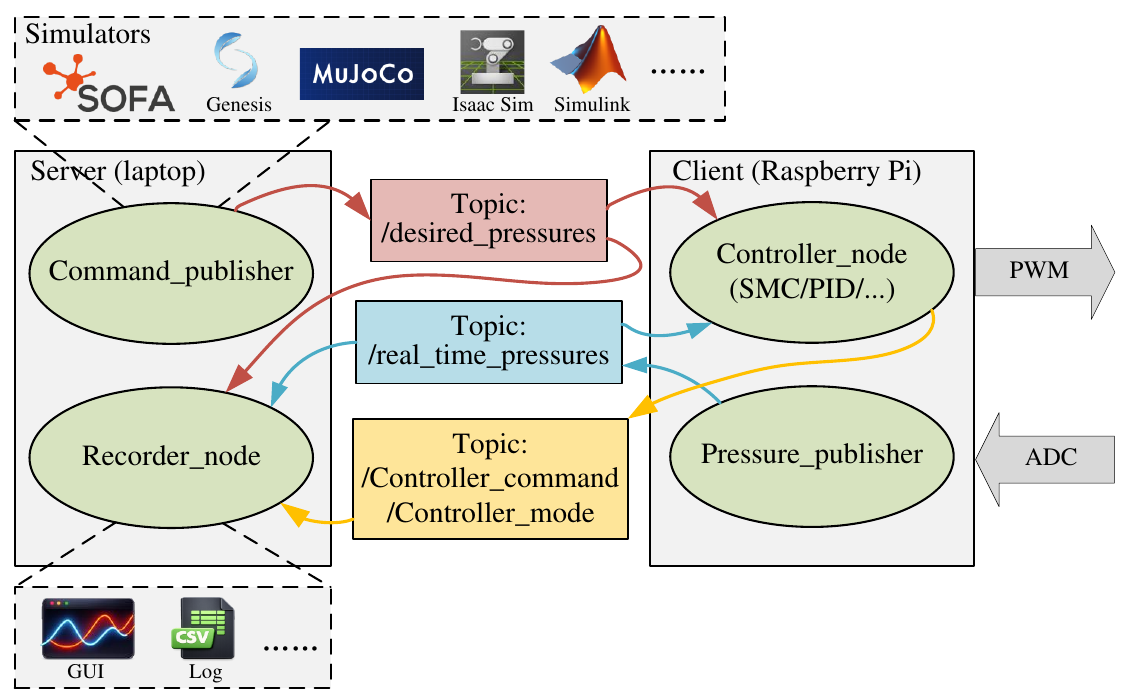}
    \caption{ROS2-based software architecture of BiPneu.}
\label{fig:supp_Software}
\end{figure}

\paragraph{Client-side (embedded execution).}
On the client side, the embedded computer (Raspberry Pi) runs a \texttt{Pressure\_publisher} node that acquires pressure measurements from the pressure sensors via the ADC Hat and publishes real-time data on the topic \texttt{/real\_time\_pressures}. In parallel, the \texttt{Controller\_node} subscribes to the desired pressure references from \texttt{/desired\_pressures} and the measured pressures from \texttt{/real\_time\_pressures}. Based on these inputs, the node executes a local pressure controller (e.g., PID or DM-SMC; see Sec.~\ref{sec:DM-SMC}) and outputs PWM signals to the PWM Hat for valve actuation. In addition, the \texttt{Controller\_node} publishes the computed control commands and the current operating mode on the topics \texttt{/controller\_command} and \texttt{/controller\_mode}, respectively. These topics are used for real-time monitoring, debugging, and synchronized data logging.

\paragraph{Server-side (host computer).}
On the server side, a \texttt{Command\_publisher} node generates desired pressure trajectories or task-level pressure commands and publishes them on the topic \texttt{/desired\_pressures}. A \texttt{Recorder\_node} subscribes to all relevant topics, including pressure measurements, control commands, and operating modes, to support real-time GUI visualization and synchronized CSV logging for offline analysis.

\subsection{Illustration of the DM-SMC Mechanism}
\label{sec:supp_dmsmc}

\begin{figure}[h] \centering \includegraphics[width=0.8\columnwidth]{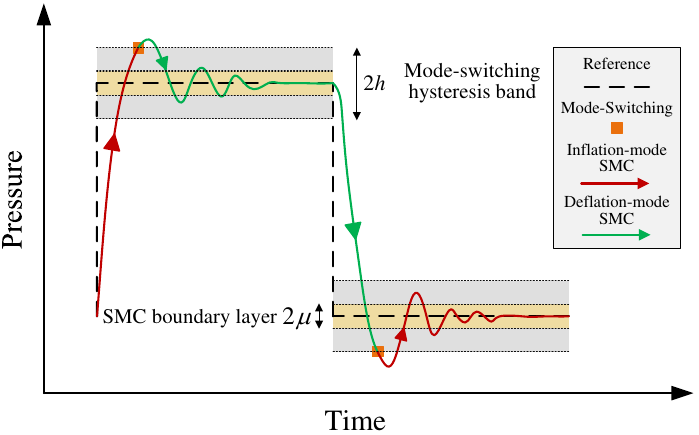} \caption{Conceptual illustration of the proposed DM-SMC.}\label{fig:supp_dm_smc_concept} \end{figure}

Fig.~\ref{fig:supp_dm_smc_concept} provides a conceptual illustration of the proposed DM-SMC on a representative step response. The control strategy consists of two coordinated parts: a discrete mode-selection layer and a continuous sliding-mode regulation layer.

First, the discrete mode $m\in\{0,1\}$ is determined by the hysteresis-based switching rule in Eq.~\eqref{eq:mode_selection}. As illustrated in Fig.~\ref{fig:supp_dm_smc_concept}, the reference pressure is surrounded by an outer hysteresis band of width $2h$. The mode variable $m$ is updated only when the tracking error defined in Eq.~\eqref{eq:error} causes the pressure to leave this band, thereby preventing rapid switching near the setpoint.

Once the discrete mode is fixed, the controller applies the mode-specific SMC law in Eq.~\eqref{eq:x_star}. The sliding surface is defined by Eq.~\eqref{eq:sliding_surface}, and its convergence is enforced through the reaching law in Eq.~\eqref{eq:reaching_law}. As shown in Fig.~\ref{fig:supp_dm_smc_concept}, this continuous regulation acts within an inner boundary layer of thickness $2\mu$, which reduces chattering and refines tracking performance.

Overall, DM-SMC combines the hysteresis-based mode-selection rule in Eq.~\eqref{eq:mode_selection} with the mode-specific SMC law in Eq.~\eqref{eq:x_star} to coordinate discrete switching and continuous valve actuation.

\vspace{-5mm}
\subsection{Simulation and Experiment Results with Detailed Standard Deviations}
The following tables report the corresponding versions with standard deviations added  (shown in parentheses), and each supplementary table matches its counterpart in the main manuscript one by one: Table~\ref{tab:supp_Sim_Step} corresponds to Table~\ref{tab:Sim_Step}, Table~\ref{tab:supp_Sim_Sine} corresponds to Table~\ref{tab:Sim_Sine}, Table~\ref{tab:supp_Exp_Step_Merged} corresponds to Table~\ref{tab:Exp_Step_Merged}, and Table~\ref{tab:supp_Exp_Sine_Merged} corresponds to Table~\ref{tab:Exp_Sine_Merged}.
\label{sec:supp_table_std}

\begin{table}[h]
    \centering
    \caption{Multi-step reference-tracking performance in simulation: PID, NMPC, MI-NMPC and DM-SMC.}
    \label{tab:supp_Sim_Step}  
    \renewcommand{\arraystretch}{1.05}%
    \setlength{\tabcolsep}{3pt}
    \resizebox{1.0\columnwidth}{!}{%
    \begin{tabular}{@{}>{\raggedright\arraybackslash}p{2.5cm}|cccc@{}}
        \hline
        Metrics & PID & NMPC & MI-NMPC & DM-SMC \\
        \hline
        $e_{\mathrm{ss}}$ [kPa]      & 1.67 (0.74)   & 1.27 (0.55)   & \textbf{0.73 (0.30)}    & 0.80 (0.43)    \\
        AE [kPa]                     & 2.48 (0.55)   & 1.91 (0.35)   & \textbf{1.49 (0.29)}    & 1.59 (0.44)    \\
        ITAE [kPa$\cdot$s$^2$]       & 22.15 (8.40)  & 15.00 (4.75)  & \textbf{9.58 (3.26)}    & 10.79 (5.19)   \\
        PWM-E [\%$\cdot$s]           & 145.60 (11.81) & 140.48 (36.67) & \textbf{135.58 (35.11)}  & 137.11 (37.40)  \\
        Switches                     & 9.08 (8.47)   & 6.08 (4.57)   & 9.08 (6.01)            & \textbf{0.83 (0.47)}    \\
        CT (PC) [ms]                 & \textbf{0.002}   & 21.528 & 355.534          & 0.060   \\
        CT (Embedded) [ms]           & \textbf{0.018}   & 270.819 & 3595.999        & 0.692   \\
        \hline
    \end{tabular}%
    }
    \begin{flushleft}
        \footnotesize \textbf{Bold} numbers represent the best results among all controllers for each metric. Values in parentheses indicate standard deviations.
        \footnotesize CT neglects the std since they almost has 0 std.
    \end{flushleft}
\end{table}

\begin{table}[h]
    \centering
    \caption{Sinusoidal reference tracking performance in simulation: PID, NMPC, MI-NMPC, and DM-SMC.}
    \label{tab:supp_Sim_Sine}  
    \vspace{-1mm}
    \renewcommand{\arraystretch}{1.05}%
    \resizebox{\columnwidth}{!}{%
    \begin{tabular}{@{}%
        >{\centering\arraybackslash}p{0.8cm}  %
        >{\centering\arraybackslash}p{2.2cm} | %
        cccc@{}}
        \hline
        $f$ [Hz] & Metrics & PID & NMPC & MI-NMPC & DM-SMC \\
        \hline
        \multirow{4}{*}{0.10}
        & AE [kPa]           & 1.64 (0.14)   & 1.14 (0.15)   & 0.78 (0.02)   & \textbf{0.72 (0.05)}   \\
        & $\max |e|$ [kPa]   & 7.19 (0.54)   & 13.48 (3.00)  & \textbf{3.89 (0.27)}  & 5.09 (0.14)   \\
        & PWM-E [\%$\cdot$s] & 268.80 (7.94) & 258.15 (4.31) & 256.11 (2.52) & \textbf{248.89 (0.39)} \\
        & Switches           & 14.67 (6.60)  & 10.00 (2.83)  & 37.00 (2.49)  & \textbf{2.00 (0.00)}   \\
        \hline
        \multirow{4}{*}{0.25}
        & AE [kPa]           & 1.61 (0.04)   & 1.33 (0.02)   & 1.20 (0.03)   & \textbf{0.78 (0.03)}   \\
        & $\max |e|$ [kPa]   & 6.72 (0.24)   & 6.52 (0.30)   & \textbf{4.71 (0.19)}  & 5.79 (0.23)   \\
        & PWM-E [\%$\cdot$s] & 119.07 (1.21) & 113.09 (2.16) & 111.71 (0.28) & \textbf{108.81 (0.13)} \\
        & Switches           & 7.00 (0.82)   & 4.67 (2.49)   & 16.67 (4.19)  & \textbf{2.00 (0.00)}   \\
        \hline
        \multirow{4}{*}{0.50}
        & AE [kPa]           & 1.80 (0.09)   & 2.17 (0.02)   & 2.18 (0.02)   & \textbf{0.86 (0.04)}   \\
        & $\max |e|$ [kPa]   & 7.68 (0.91)   & 7.31 (0.81)   & \textbf{5.52 (0.22)}  & 6.66 (0.23)   \\
        & PWM-E [\%$\cdot$s] & 70.25 (1.71)  & 66.31 (0.47)  & \textbf{65.35 (0.49)} & 65.59 (0.90)  \\
        & Switches           & 3.33 (0.94)   & 2.67 (0.94)   & 9.33 (2.05)   & \textbf{2.00 (0.00)}   \\
        \hline
    \end{tabular}%
    }
\end{table}

\begin{table}[h]
    \centering
    \caption{Experimental multi-step reference tracking performance under fixed-volume and varying-volume loads: PID vs. DM-SMC.}
    \label{tab:supp_Exp_Step_Merged}
    \renewcommand{\arraystretch}{1.2}%
    \resizebox{0.95\columnwidth}{!}{%
    \begin{tabular}{@{}>{\centering\arraybackslash}p{2.8cm}|cc|cc@{}}
        \hline
        \multirow{2}{*}{Metrics}
        & \multicolumn{2}{c|}{Fixed-volume}
        & \multicolumn{2}{c@{}}{Varying-volume} \\
        \cline{2-5}
        & PID & DM-SMC & PID & DM-SMC \\
        \hline
        $e_{\mathrm{ss}}$ [kPa]         & 1.52 (1.57)   & \textbf{1.44 (1.70)}   & 1.62 (1.19)   & \textbf{1.43 (1.46)} \\
        AE [kPa]                        & 2.87 (3.59)   & \textbf{2.53 (2.98)}   & 2.72 (1.62)   & \textbf{2.59 (2.64)} \\
        ITAE [kPa$\cdot$s$^2$]          & 57.99 (35.82) & \textbf{28.18 (27.60)} & \textbf{26.41 (16.06)} & 27.27 (23.88) \\
        PWM-E [\%$\cdot$s]              & 132.93 (15.64) & \textbf{126.89 (13.71)} & 134.30 (18.27) & \textbf{129.01 (13.16)} \\
        Switches                        & 2.92 (2.29)   & \textbf{1.42 (1.66)}   & 2.75 (2.69)   & \textbf{1.33 (0.72)} \\
        \hline
    \end{tabular}%
    }
\end{table}

\begin{table}[h!]
    \centering
    \caption{Experimental sinusoidal reference tracking performance under fixed-volume and varying-volume loads: PID vs. DM-SMC.}
    \label{tab:supp_Exp_Sine_Merged}
    \renewcommand{\arraystretch}{1.2}%
    \resizebox{0.95\columnwidth}{!}{%
    \begin{tabular}{@{}>{\centering\arraybackslash}p{0.9cm}>{\centering\arraybackslash}p{2.0cm}|cc|cc@{}}
        \hline
        \multicolumn{1}{c}{\multirow{2}{*}{$f$ [Hz]}} 
        & \multicolumn{1}{c|}{\multirow{2}{*}{Metrics}}
        & \multicolumn{2}{c|}{Fixed-volume}
        & \multicolumn{2}{c@{}}{Varying-volume} \\
        \cline{3-6}
        & & PID & DM-SMC & PID & DM-SMC \\
        \hline
        \multirow{4}{*}{0.10}
        & AE [kPa]               & \textbf{4.00 (0.17)} & 4.01 (0.29) & \textbf{3.62 (0.15)} & 4.19 (0.04) \\
        & $\max |e|$ [kPa]       & \textbf{13.88 (1.57)} & 18.73 (0.60) & \textbf{14.15 (1.06)} & 15.21 (0.92) \\
        & PWM-E [\%$\cdot$s]     & 250.07 (0.45) & \textbf{247.39 (0.53)} & \textbf{251.09 (2.10)} & 256.23 (0.27) \\
        & Switches               & \textbf{4.00 (1.63)} & 4.67 (0.94) & 3.33 (0.94) & \textbf{2.00 (0.00)} \\
        \hline
        \multirow{4}{*}{0.25}
        & AE [kPa]               & 4.38 (0.16) & \textbf{2.97 (0.28)} & 3.60 (0.47) & \textbf{2.16 (0.06)} \\
        & $\max |e|$ [kPa]       & 16.80 (1.77) & \textbf{15.73 (2.33)} & 19.49 (3.20) & \textbf{8.92 (0.45)} \\
        & PWM-E [\%$\cdot$s]     & 107.07 (1.73) & \textbf{104.11 (0.56)} & 108.55 (1.41) & \textbf{107.41 (0.30)} \\
        & Switches               & 2.67 (0.94) & \textbf{2.00 (0.00)} & 2.67 (0.94) & \textbf{2.00 (0.00)} \\
        \hline
        \multirow{4}{*}{0.50}
        & AE [kPa]               & 6.57 (0.93) & \textbf{4.23 (0.88)} & 7.17 (1.89) & \textbf{3.60 (0.44)} \\
        & $\max |e|$ [kPa]       & 29.54 (5.34) & \textbf{16.10 (2.39)} & 40.43 (9.00) & \textbf{14.37 (2.07)} \\
        & PWM-E [\%$\cdot$s]     & 62.16 (1.86) & \textbf{57.97 (1.20)} & 65.16 (4.00) & \textbf{60.45 (0.30)} \\
        & Switches               & 3.33 (0.94) & \textbf{2.00 (0.00)} & 2.00 (0.00) & 2.00 (0.00) \\
        \hline
    \end{tabular}%
    }
\end{table}

\subsection{Multi-level Steps Responses for Both Simulations and Experiments}
\label{sec:supp_steps}

\begin{figure}[!hbp]
\centering
\includegraphics[width=0.94\columnwidth]{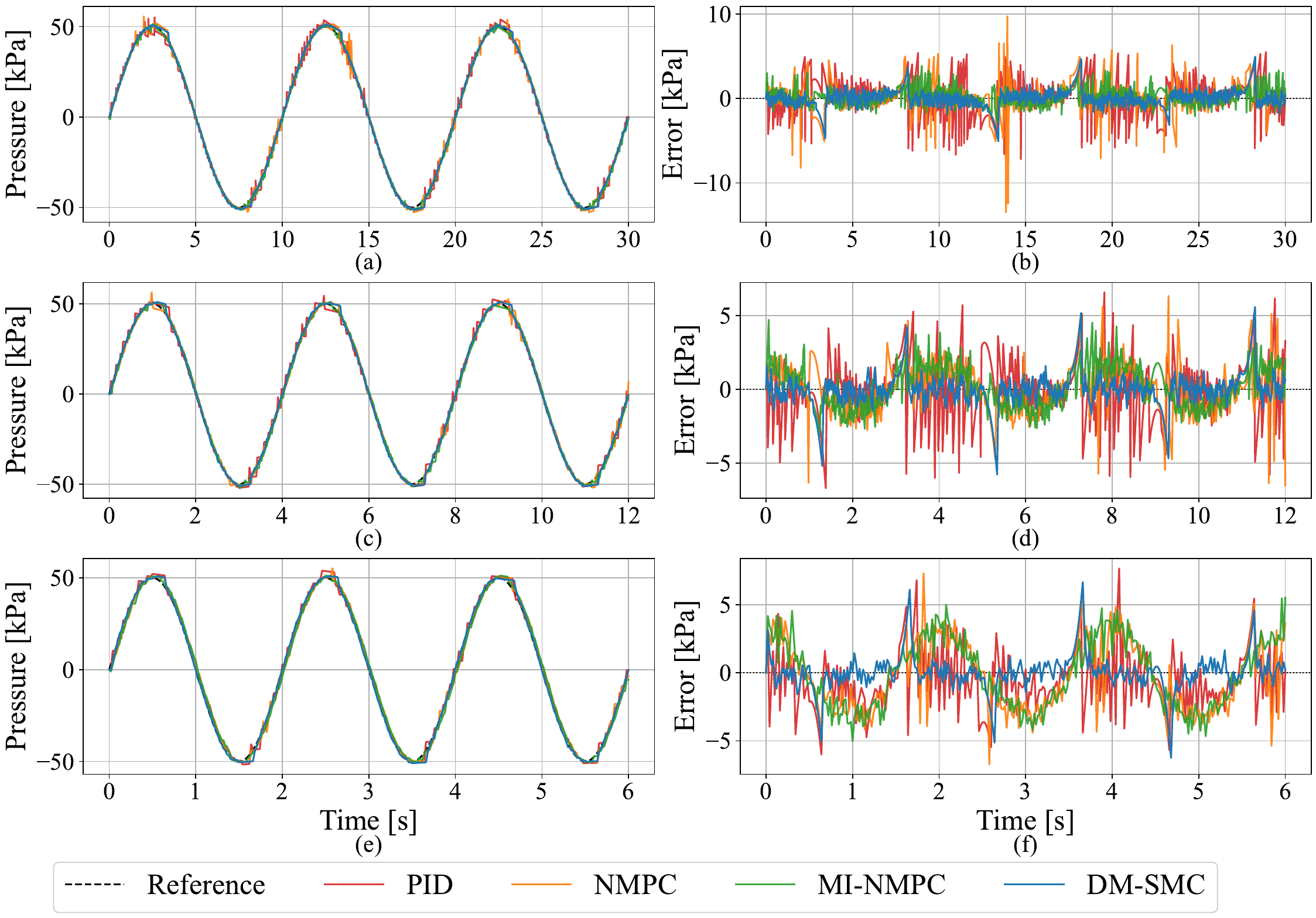}
\caption{Simulation results of sinusoidal reference tracking performance for different controllers: (a) pressure trajectories under the reference and four controllers; (b) corresponding tracking errors.}
\label{fig:supp_Sim_Sine}
\vspace{-3mm}
\end{figure}

\begin{figure}[!t]
\centering
\includegraphics[width=0.94\columnwidth]{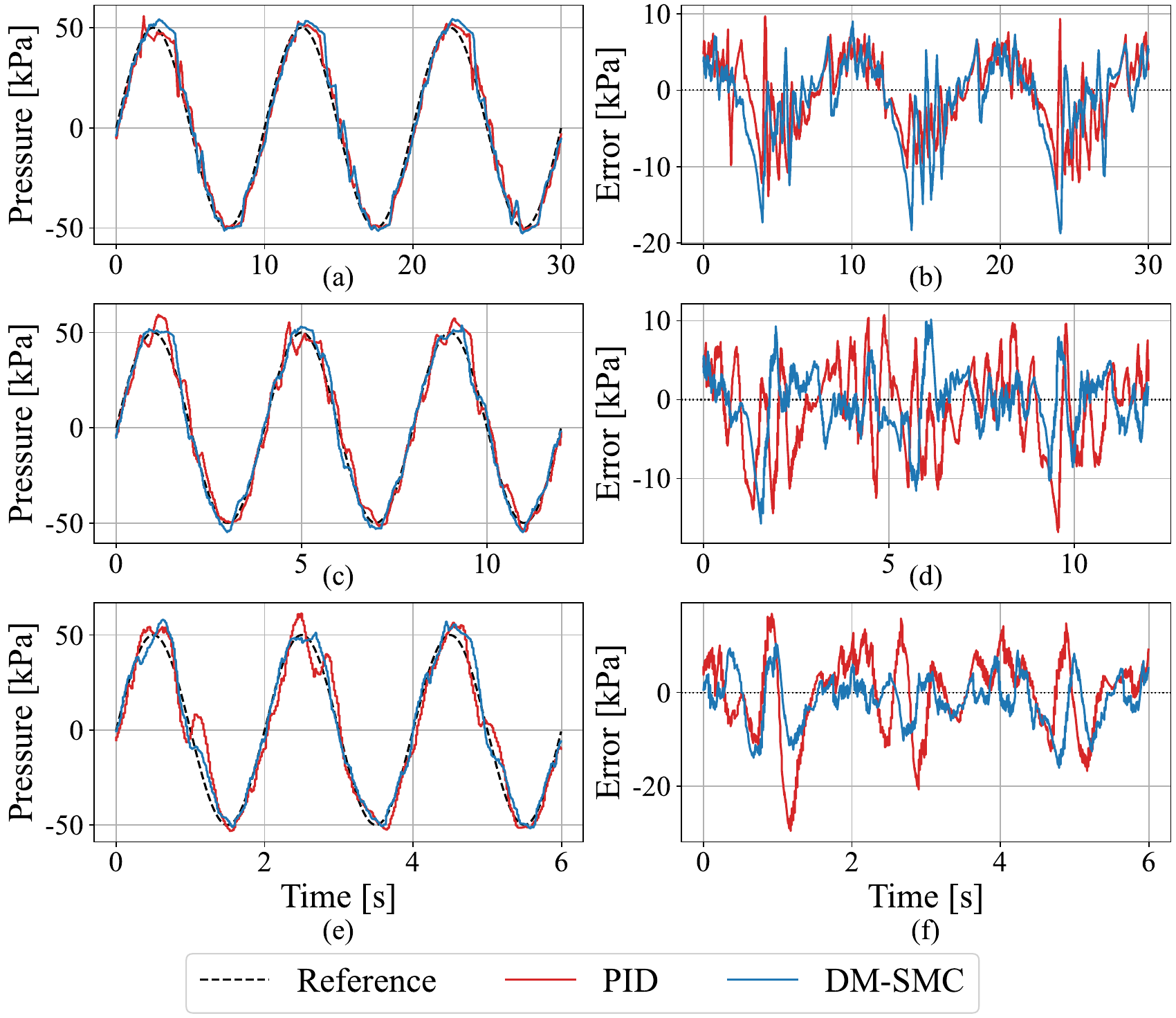}
\caption{Experimental sinusoidal reference tracking performance of PID and DM-SMC at three frequencies: (a)--(b) pressure trajectories and tracking errors at 0.1 Hz; (c)--(d) at 0.25 Hz; (e)--(f) at 0.5 Hz.}
\label{fig:supp_Exp_Sine}
\vspace{-4mm}
\end{figure}

The relatively larger error at $0.1$~Hz is likely due to the higher mismatch of the identified mapping $\bar{x}(u)$ under slowly varying references, which leads to the observed non-monotonic trend across frequencies.

\vspace{-2mm}

\subsection{Tracking Responses Under Varying-Volume Bellow Load}
\label{sec:supp_bellow}

This section provides the measured step- and sine-tracking responses under a varying-volume bellow-based load, which are omitted from the main text due to space limitations.

\begin{figure}[!b]
\centering
\includegraphics[width=0.96\columnwidth]{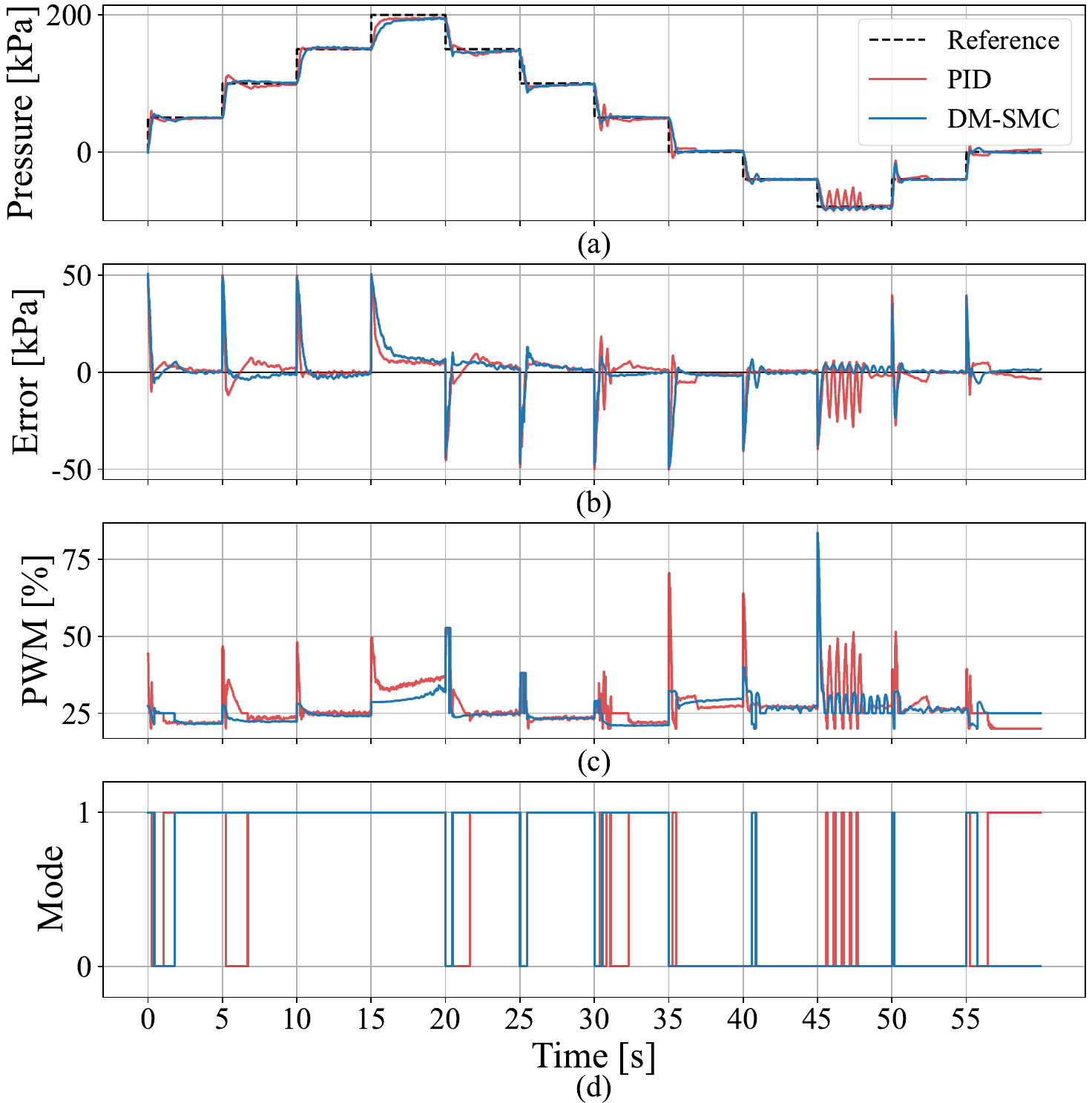}
\vspace{-3mm}
\caption{Experimental results of multi-step reference-tracking using PID and DM-SMC to control a bellow-like soft actuator acting as a volume-varying load. (a) pressure trajectories of PID, DM-SMC and reference; (b) corresponding tracking errors; (c) PWM signals; (d) valve mode sequence.}
\label{fig:supp_Exp_Step_Bellow}
\vspace{-4mm}
\end{figure}

\FloatBarrier

\begin{figure}[!t]
\centering
\includegraphics[width=0.94\columnwidth]{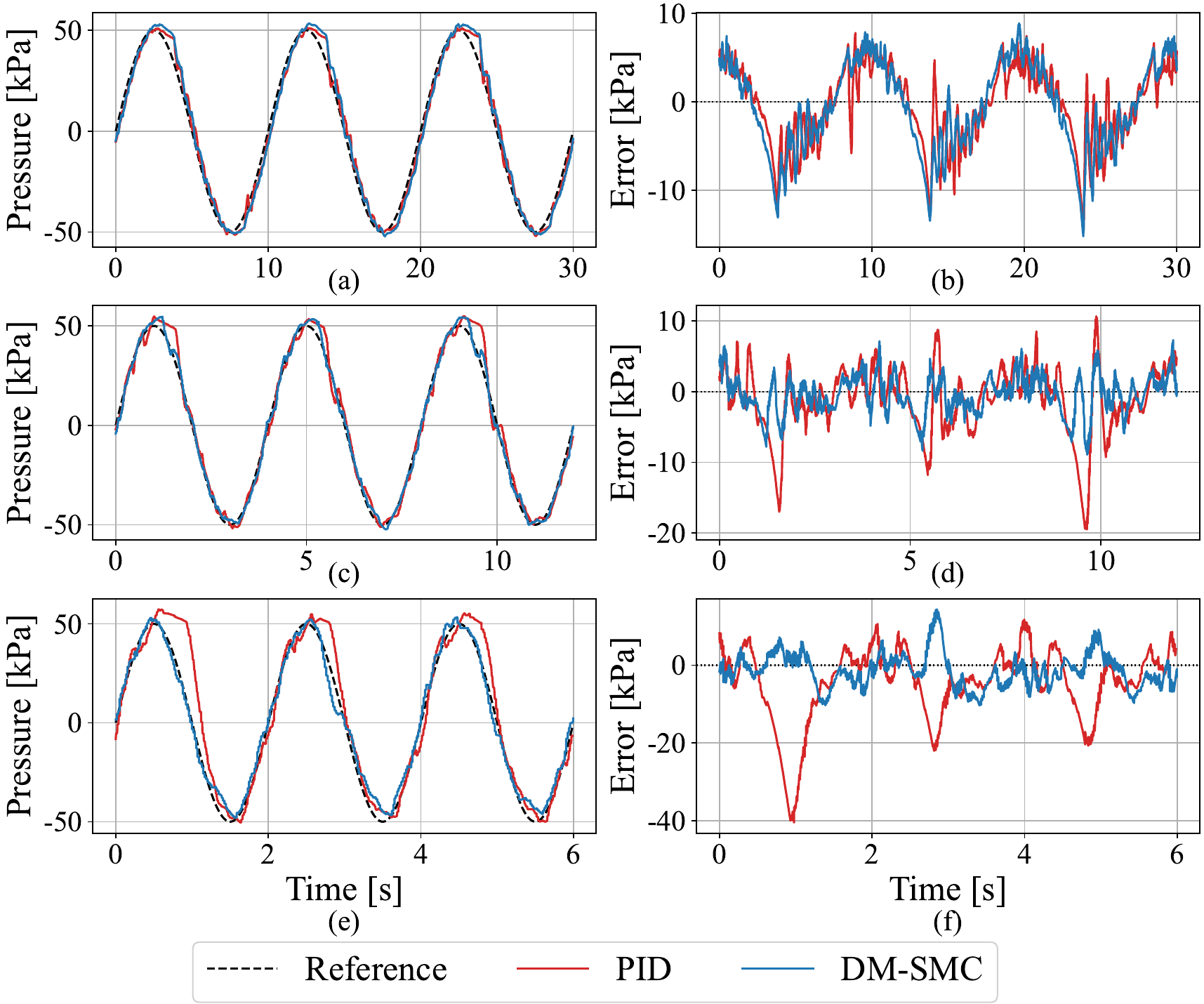}
\caption{Experimental results of sinusoidal reference-tracking of a bellow-like soft actuator using PID and DM-SMC at three frequencies: (a)--(b) pressure trajectories and tracking errors at 0.1 Hz; (c)--(d) at 0.25 Hz; (e)--(f) at 0.5 Hz.}
\label{fig:supp_Exp_Sine_Bellow}
\vspace{-3mm}
\end{figure}

\subsection{FEM-Based Teleoperation Validation}
\label{sec:supp_teleop}

\begin{figure}[!hbp]
\centering
\includegraphics[width=0.68\columnwidth]{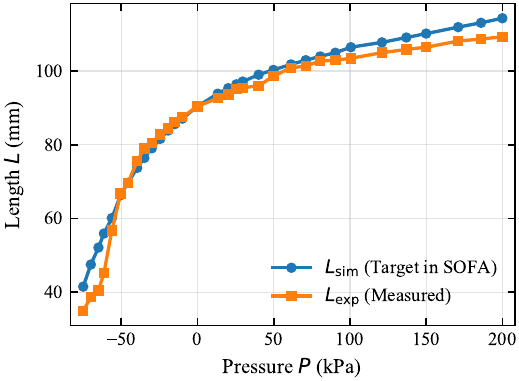}
\caption{FEM-based teleoperation results: SOFA target bellow length $L_{\text{sim}}$ versus measured length $L_{\text{exp}}$ across pressure $P$.}
\label{fig:supp_SOFA_results}
\vspace{-3mm}
\end{figure}

\FloatBarrier

\addcontentsline{toc}{section}{Supplementary Material}

\end{document}